\documentclass[journal=jacsat,manuscript=article]{achemso}

\usepackage[version=3]{mhchem} 
\usepackage{orcidlink}
\usepackage[T1]{fontenc}
\usepackage{amsmath}
\usepackage{amsfonts}
\usepackage{placeins}
\usepackage{hyperref}
\usepackage{graphicx}
\usepackage{caption, subfig}
\usepackage[frozencache,cachedir=.]{minted}
\definecolor{customblue}{RGB}{42, 130, 218}
\definecolor{customred}{RGB}{231, 76, 60}
\definecolor{BlueGreen}{RGB}{0, 128, 128}
\setminted{
    frame=single,           
    fontsize=\footnotesize,   
    bgcolor=white,          
    breaklines=true,        
    style=pastie,
    linenos=false,           
    escapeinside=||,
}
\graphicspath{{figures/}}

\usepackage{adjustbox,tikz}
\usetikzlibrary{scopes,quotes,positioning,matrix,calc,arrows.meta, fit}
\usetikzlibrary{graphs,shapes.symbols,shapes.misc,shapes.arrows}
\usetikzlibrary{3d,decorations.text,backgrounds}

\usepackage{algorithm}
\usepackage{algorithmic}
\usepackage{multirow}
\usepackage{booktabs} 
\usepackage{array}

\author{Siddhant Dutta\orcidlink{0009-0000-5120-7114}}
\email{forsomethingnewsid@gmail.com}
\affiliation[Dwarkadas J. Sanghvi College of Engineering]
{SVKM's Dwarkadas J. Sanghvi College of Engineering, Mumbai 400056, India}

\author{Iago Leal de Freitas\orcidlink{0009-0001-6813-5863}}
\email{ilealdef@purdue.edu}
\affiliation[Purdue University]
{Davidson School of Chemical Engineering, Purdue University, West Lafayette, IN 47907, USA}

\author{Pedro Maciel Xavier\orcidlink{0000-0002-4678-4942}}
\email{pmacielx@purdue.edu}
\affiliation[Purdue University]
{Davidson School of Chemical Engineering, Purdue University, West Lafayette, IN 47907, USA}
\alsoaffiliation[Federal University of Rio de Janeiro]
{PESC/COPPE, Federal University of Rio de Janeiro, RJ, 21941-853, Brazil}

\author{Claudio Miceli de Farias\orcidlink{0000-0002-1927-7398}}
\email{cmicelifarias@cos.ufrj.br}
\affiliation[Federal University of Rio de Janeiro]
{PESC/COPPE, Federal University of Rio de Janeiro, RJ, 21941-853, Brazil}

\author{David E. Bernal Neira\orcidlink{0000-0002-8308-5016}}
\email{dbernaln@purdue.edu}
\affiliation[Purdue University]
{Davidson School of Chemical Engineering, Purdue University, West Lafayette, IN 47907, USA}

\title[Federated Learning in Chemical Engineering]
  {Federated Learning in Chemical Engineering: A Tutorial on a Framework for Privacy-Preserving Collaboration Across Distributed Data Sources}

\abbreviations{FL, ML, FedAvg, FedOpt, IID, FedProx, FlChain}

\begin{document}







\begin{abstract}
Federated Learning (FL) is a decentralized machine learning approach that has gained attention for its potential to enable collaborative model training across clients while protecting data privacy, making it an attractive solution for the chemical industry.
This work aims to provide the chemical engineering community with an accessible introduction to the discipline.
Supported by a hands-on tutorial and a comprehensive collection of examples, it explores the application of FL in tasks such as manufacturing optimization, multimodal data integration, and drug discovery while addressing the unique challenges of protecting proprietary information and managing distributed datasets.
The tutorial was built using key frameworks such as \texttt{Flower} and \texttt{TensorFlow Federated} and was designed to provide chemical engineers with the right tools to adopt FL for their specific needs.
We compare the performance of FL against centralized learning across three different datasets relevant to chemical engineering applications, demonstrating that FL will often maintain or improve classification performance, particularly for complex and heterogeneous data.
We conclude with an outlook on the open challenges in federated learning to be tackled and current approaches designed to remediate and improve this framework.
\end{abstract}

\section{Introduction}

Artificial intelligence has gained significant prominence with recent market innovations and the development of powerful hardware.
In traditional machine learning (ML), the efficiency and accuracy of the models depend on the computational power and training data of a centralized server. 
Therefore, several factors have been decisive in its evolution, such as the availability of a vast amount of data collected in recent years (Big Data), the evolution of computational power in devices, and also the emergence of deep learning (DL) models \cite{VMothukuri}.

In traditional ML, user data is stored on the central server and used for training and testing processes to develop comprehensive models.
Centralized ML approaches are generally associated with different challenges, including computational power and time, and most importantly, security and privacy regarding user data \cite{VMothukuri}.

Although ML models are capable of providing knowledge and computational power at a specific cost, concerns about data privacy and confidentiality remain unresolved \cite{Caviglione, KeidelR2018, Cabaj2018-1, Lyu2019, Cabaj2018-2}.
Federated Learning (FL), proposed by \citet{mcmahan2023communicationefficientlearningdeepnetworks}, has recently emerged as a technological solution to address these problems and is one of the growing fields in ML.
Its appeal comes from its concept of decentralized data, and it provides security and privacy features that promise to comply with the emerging laws on user data protection \cite{Anonymous_2013, european_commission_regulation_2016}.

Federated Learning (FL) is a distributed ML approach that enables various devices to collaboratively train a global model without sharing raw data.
Instead, local devices compute model updates that are aggregated to form a global model.
This decentralized method helps maintain user privacy and data security by preventing the centralization of sensitive information. 

\begin{figure}[tb]
\centering
\begin{tikzpicture}[]

\colorlet{client1}{purple!90}
\colorlet{client2}{blue!40!black}
\colorlet{client3}{ green!60!black}

\pgfdeclareimage[width=25mm, height=25mm]{factory}{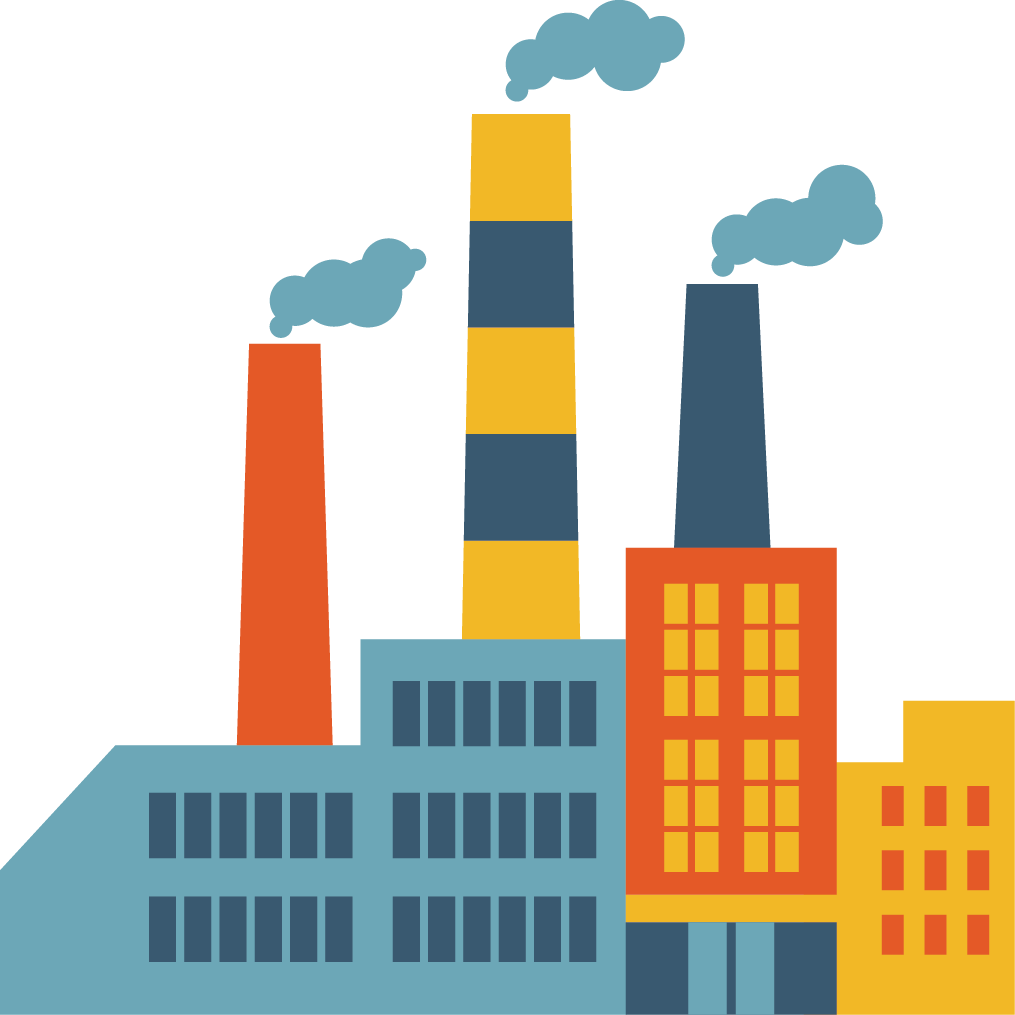}
\pgfdeclareimage[width=25mm, height=25mm]{factory2}{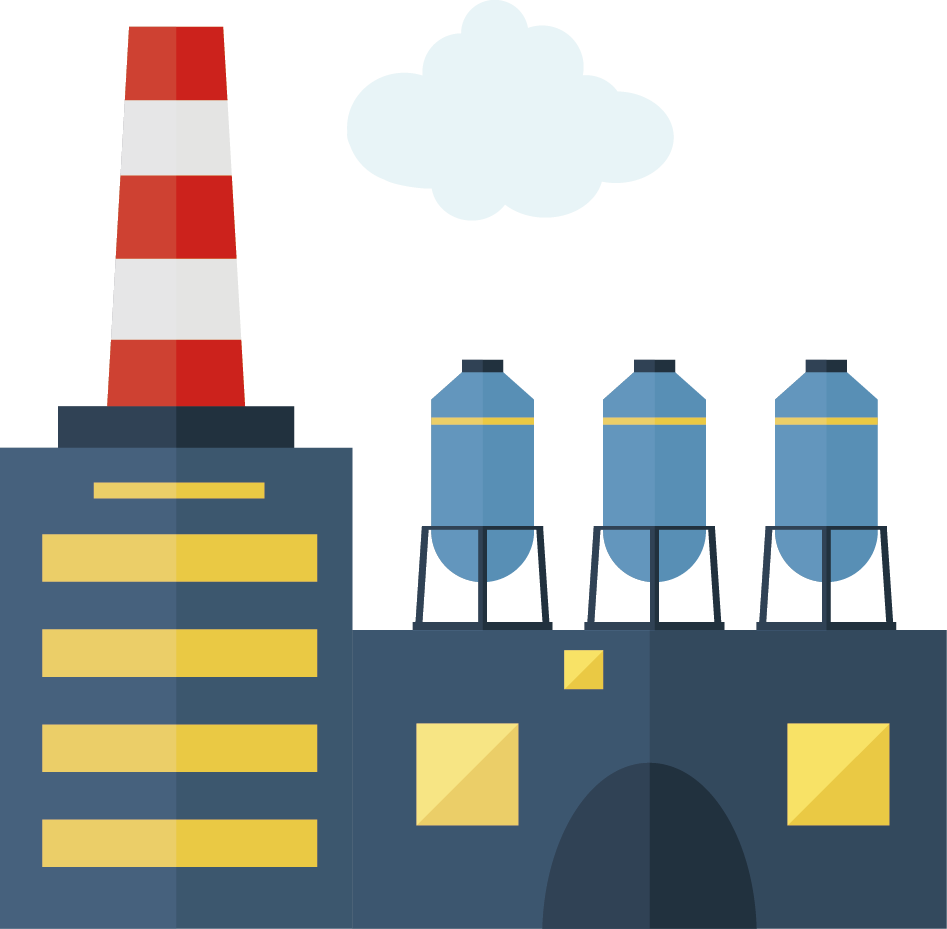}

\pgfdeclarelayer{background}
\pgfdeclarelayer{background2}
\pgfdeclarelayer{foreground}
\pgfsetlayers{background,background2,main,foreground}

    \tikzset{
        data1/.pic= {
            \begin{scope}[scale=0.625, pic actions]
                \path[fill=red!70]  (0, 0) rectangle (1, 1);
                \path[fill=blue!80!black] (1, 0) rectangle (2, 1);
                \path[fill=green!40] (2, 0) rectangle (3, 1);
                \path[fill=yellow] (3, 0) rectangle (4, 1);
            \end{scope}
        },
        data2/.pic= {
            \begin{scope}[scale=0.625, pic actions]
               \path[fill=red!70]  (0, 0) rectangle (0.6, 1);
               \path[fill=blue!80!black] (0.6, 0) rectangle (2.1, 1);
               \path[fill=green!40] (2.1, 0) rectangle (2.8, 1);
               \path[fill=yellow] (2.8, 0) rectangle (4, 1);
            \end{scope}
        },
        neuralnetwork/.pic={
            \pgfkeys{/pic/args/.default={2,3,3}} 
            \def\layers{{#1}}

            \begin{scope}[
              state/.style    = { circle, minimum width = 0.3cm, draw = gray!40, very thick, outer sep = 1mm, pic actions}, 
              action/.style   = {
                opacity = 1,
                thin,
                black!80,
                ->,
                -{Kite[length=4pt,width=2.5pt,inset=1.5pt]},
              },
            ]
              \pgfmathsetmacro{\nLayers}{dim(\layers)}
              \expandafter\pgfmathmax\layers{}\edef\maxStates{\pgfmathresult}

              \foreach \t in {\nLayers,...,1} {
                \pgfmathsetmacro{\nStates}{\layers[\t-1]}

                \foreach \s in {\nStates,...,1} {
                  \node[state, outer sep = 2] (-q\t\s) at ({1cm*(\t-2)}, {0.6cm*(\s - (\nStates+1)/2)}) {};

                  \ifnum \t<\nLayers
                    \pgfmathtruncatemacro{\next}{\t + 1}
                    \pgfmathsetmacro{\nStatesNext}{\layers[\t]}

                    \foreach \ny in {1,...,\nStatesNext} {
                      \draw [action] (-q\t\s) -- (-q\next\ny);
                    }
                  \fi
                }
              }
              
            \begin{pgfonlayer}{background}
              \node[rectangle, thick, fill=white!20, draw=gray!30, fit = (-q11) (-q33), inner sep = 2mm] (-box) {};
            \end{pgfonlayer}
            
            \end{scope}
        }
    }

\begin{scope}[ ]
  \node[] (U1){\pgfuseimage{factory}};
  \node[right = 3.5cm of U1] (U2) {\pgfuseimage{factory2}};
  \node[right = 3.5cm of U2, minimum width=20mm, minimum height=20mm] (U3) {\huge$\ldots$};
\end{scope}

\pic[yshift=-7mm, xshift=1mm] (Data1) at (U1.south west) {data1};
\pic [below = 1.7cm of U1, opacity=0.4, fill=client1] (NN1) {neuralnetwork={3,2,3}};

\pic[yshift=-7mm, xshift=1mm] (Data2) at (U2.south west) {data2};
\pic [below = 1.7cm of U2, fill=client2] (NN2) {neuralnetwork={3,2,3}};

\begin{scope}[ ]
    \node[rectangle, thick, draw=blue!30!black, fill=blue!20, fill opacity = 0.2, minimum width=4cm, minimum height=2cm, below = 4.5cm of U2] (server) {};
    
    \node[below = 1mm of server.center] {\Large Server};
    \node[above = -0.1mm of server.center, draw=brown!60!black, thick, rounded corners] (agg) {\tiny $\theta^{(t+1)} \leftarrow \mathrm{Aggregate}\left(\theta_i^{(t)}\right)$};
\end{scope}

\begin{scope}[action/.style = {
    thick,
    -{Latex[length=4pt,width=2.5pt,inset=1.5pt]},
    },
]
    \draw[->, action, color=client1]
        (NN1-box.-100) |- 
        node [near start, left] {$\theta_1^{t}$}
        (server.190);
    \path[postaction={transform canvas={xshift=+1mm},draw, action}, color=client2]
        (NN2-box) --
        node [midway, right, xshift=1mm] {$\theta_2^{t}$}
        (server);
    \draw[->, action, color=client3]
        (U3.-80) |-
        node [near start, right] {$\theta_3^{t}$}
        (server.-10);

    \draw[->, action, brown!60!black]
        (agg) -|
        node [near end, right, yshift=-3mm] {$\theta^{(t+1)}$}
        (NN1-box.-80);
    \path[postaction={transform canvas={xshift=-1mm},draw, action}, brown!60!black]
        (agg) --
        node [midway, left, xshift=-1mm] {$\theta^{(t+1)}$}
        (NN2-box);
    \draw[->, action, brown!60!black]
        (agg) -|
        node [near end, left, yshift=-4mm] {$\theta^{(t+1)}$}
        (U3.-100);
    \end{scope}
\end{tikzpicture}
\caption{Each client trains their respective models on local data.
After each training round, they transmit their local weights $\theta_k^{(t)}$ to a central server for aggregation. The new global model $\theta^{(t+1)}$ is then distributed to all clients.
They substitute it as their current model and proceed with another training round.
This way, the clients have access to global information but never directly observe each other's datasets%
~\cite{img_factory1,img_factory2}.
}
\label{fig:workflow}
\end{figure}

FL presents several potential scenarios across industries, each leveraging its decentralized architecture to address specific challenges.
In healthcare, it enables hospitals and research institutions to collaboratively train diagnostic models without exposing sensitive patient data, improving early disease detection~\cite{nguyen2022federated, bhatia2024federatedhierarchicaltensornetworks}.
This application has stringent privacy requirements as the training data are patient diagnoses and images~\cite{zhang2022homomorphic, ku2022privacy, lessage2024secure}
In finance, it allows banks to develop fraud detection algorithms by combining insights from multiple sources without compromising customer privacy~\cite{innan2024qfnn}.
For autonomous vehicles, FL facilitates the sharing of driving data between manufacturers to improve self-driving algorithms while protecting proprietary information.
In industrial sectors like manufacturing and energy, predictive maintenance is supported by combining data from various plants without centralizing it~\cite{JAVEED2024577} \cite{nguyen2021federated} \cite{qu2020blockchained} \cite{otoum2022federated} \cite{gao2022secure}. 
Figure~\ref{fig:workflow} illustrates this decentralized procedure,
where different manufacturers only have access to the trained model.

Not sharing data can be particularly relevant in the chemical industry, as companies often deal with sensitive data related to proprietary chemical formulas, production processes, and safety protocols.
Using FL, organizations can build robust predictive models for tasks such as material discovery, process optimization, and chemical safety assessments while maintaining data privacy. 


Some works applying FL to problems relevant to chemical engineering are presented here.
\citet{heyndrickx2023melloddy} presents an application of FL on Quantitative Structure-Activity Relationship (QSAR) models.
In \citet{bassani2023federated}, the authors show an application in toxicology.
Molecular discovery has been presented as a promising application of FL by \citet{hanser2023federated}, and \citet{oldenhof2023industry} have used it to solve a challenging drug discovery problem.
As we can observe, FL fosters collaboration between companies and research institutions, accelerating innovation while ensuring compliance with stringent data protection regulations.

Some works applying FL to problems relevant to chemical engineering are presented here. For example, \citet{heyndrickx2023melloddy} demonstrate the use of FL to develop Quantitative Structure-Activity Relationship (QSAR) models, which are essential to predict molecular properties and drug efficacy. By training QSAR models in a federated setting, organizations can benefit from a larger and more diverse dataset while keeping sensitive compound information confidential. Similarly, \citet{bassani2023federated} apply FL to toxicology studies, where combining data from multiple laboratories improves the reliability of toxicity assessments without exposing proprietary experimental details.

Another compelling application is pill image classification by \citet{Bergmann_2021}, which plays a crucial role in quality control within pharmaceutical manufacturing. Given the enormous number of pill images produced in manufacturing plants, manual inspection is impractical. Furthermore, because defective pills (failures) are rare events in any single plant, pooling data from multiple facilities through FL not only improves the robustness of the model in detecting these rare defects but also allows each plant to maintain data privacy. This is critical because the release of raw images could inadvertently reveal sensitive information about proprietary manufacturing processes. Molecular discovery and drug discovery have also been investigated using FL, as evidenced by the work of \citet{hanser2023federated} and \citet{oldenhof2023industry}. These studies illustrate how federated approaches can accelerate innovation by enabling multiple organizations to collaboratively train models while complying with strict data protection regulations.

Several FL tutorials are available in the academic literature \cite{luzon_tutorial_2024} as well as the documentation of powerful frameworks that implement relevant tools for it, e.g., \texttt{TensorFlow Federated} \cite{The_TensorFlow_Federated_Authors_TensorFlow_Federated_2018} and \texttt{PyTorch Flower}~\cite{beutel2020flower,paszke2019pytorch}.  
Our contribution in this work is highlighting the potential uses of FL for Chemical Engineering.
We present the topic to an audience that might not have been exposed to it prior. We then provide several examples highlighting capabilities and applications relevant to this community. These include image classification for pill manufacturing, multi-fidelity models that combine genetic data and medical images, and drug discovery through federated graph-based learning.
Finally, we provide the direction of current research that aims to improve the framework.


\section{Federated Learning Theory}

Federated learning (FL) is an alternative to the standard centralized ML approach.
It is a distributed learning architecture in which local machines train their models independently while communicating them to a central server, which aggregates them.
In this section, we discuss the theory behind it as well as different possible model aggregation schemes.

\subsection{Centralized Learning}

Centralized learning (CL) in ML involves a central server or parameter server (PS) that aggregates data from distributed clients and conducts the entire training process centrally.
In this model, clients send their local datasets to the server, which processes the data and updates the model based on a global optimization objective.
The PS has complete access to the entire dataset, allowing it to perform accurate and efficient computations, particularly with large-scale heterogeneous data.
This approach often leads to high learning accuracy, as the centralized model optimizes its parameters across all available data at once, reducing the variance and bias seen in distributed models such as FL~\cite{elbir2022hybrid, TRUONG2021102402}.

However, centralized learning faces challenges in scalability and privacy.
The need to transmit all data to a central server often results in high communication overhead, especially in networked systems.
This can raise privacy concerns, as sensitive data is collected at a single point, making it vulnerable to breaches~\cite{TRUONG2021102402}.
Additionally, CL requires computational resources (in terms of processing power and memory) at the central node, and system failures at this point can collapse the entire model training process.
Despite these challenges, centralized learning remains a prominent strategy in cases where computational resources are plentiful, and data privacy is less of a concern.

This situation raises the question: what if a system could eliminate the need to centralize data, allowing for localized training on individual devices?
In such a system, the data would remain on local devices while still contributing to a collective model.
By aggregating the knowledge from locally trained models, similar or even better performance could be achieved than centralized systems without compromising data privacy or enduring high data transmission costs.
This concept is the foundation of FL~\cite{mcmahan2023communicationefficientlearningdeepnetworks}.
In FL, training happens directly on devices or local nodes, and only the model updates, such as weights or gradients, are shared with a central server for aggregation.
This framework enables large-scale collaborative learning across distributed environments without the need to move sensitive or vast datasets~\cite{10.1145/3533708}.

\subsection{Transitioning from Centralized to Federated Learning}

Denote the entire data set by $\mathcal{D}$ and the number of samples by $N$.
In centralized learning, the model parameters $\theta$ are optimized by minimizing a global loss function
\begin{equation}
\theta^* = \arg \min_{\theta} \frac{1}{N} \sum_{i=1}^{N} \mathcal{L}(f(x_i; \theta), y_i),
\end{equation}
where $\mathcal{L}$ represents the loss function, \(f\) is the function that represents the learning architecture, $x_i$ are the input features, and $y_i$ are the corresponding labels for sample \(i\).
In training the ML model, we aim to have \(f(x_i;\theta) \approx y_i\).
This global optimization is performed using the entire dataset $\mathcal{D}$, which is stored on a central server~\cite{li_review_2020}.

In contrast, FL divides this optimization task between $K$ clients, each with its local dataset $\mathcal{D}_k$.
The local optimization problem for client $k$ can be described as
\begin{equation}
\theta_k^* = \arg \min_{\theta_k} \frac{1}{|\mathcal{D}_k|} \sum_{i \in \mathcal{D}_k} \mathcal{L}(f(x_i; \theta_k), y_i).
\end{equation}
where \( |\mathcal{D}_k| \) is the size of client \(k\)’s dataset.

Once the local models are trained,
the clients send the necessary information to a central server
to aggregate these updates\cite{luzon_tutorial_2024}.
Depending on the aggregation strategy in use,
the clients may send
their model parameters $\theta_k$,
training gradients or any other necessary local information.
In the next section,
we discuss different aggregation schemes,
including which information they need.

\subsection{Data Distribution And Model Aggregation Techniques in Federated Learning}

Many traditional ML models assume that the samples are drawn uniformly from the same probability distribution,
a characteristic known as \emph{independent and identically distributed} (IID) data,
because it simplifies the mathematical models and ensures consistent statistical properties across the dataset~\cite{alhamoud2024fedmedicl}.
For example, when training a model on a large, centralized dataset, the data points can be shuffled and divided into mini-batches without concern for introducing bias due to differences in data distributions.
IID data environments ensure that the model can generalize effectively because all clients contribute equally to the learning process, and the model is exposed to uniformly distributed samples.

However, in FL, data is often non-IID due to the decentralized nature of data collection across multiple clients, such as mobile devices or institutions, each having data that reflects local user behavior or environment.
Non-IID data introduces heterogeneity, where different clients may possess vastly different distributions of data, leading to challenges in model convergence and training stability~\cite{alhamoud2024fedmedicl}.

For instance, when the data comes from manufacturing plants, the inherent differences in their production schemes imply different data distributions.
This disparity can cause models to overfit to local data and fail to generalize across the system~\cite{alhamoud2024fedmedicl}.
Handling non-IID data is a significant challenge in FL, as it complicates optimization and can lead to unfair model performance among clients.
Therefore, there are several methods to aggregate the local updates from clients in FL. The choice of aggregation technique can significantly impact the performance and efficiency of the FL system~\cite{iyer2024review}.
Below are some common aggregation techniques.

Federated Averaging (\textit{FedAvg})~\cite{mcmahan2023communicationefficientlearningdeepnetworks} is the most widely used aggregation technique in FL.
In \textit{FedAvg}, each client \(k\), of a total of \(K\) clients, trains its model locally using its dataset \(\mathcal{D}_k\), i.e., find the parameters \(\theta_k^{(t)}\) that minimize loss given its local data in the FL iteration \(t\).
The server aggregates these model updates by computing a weighted average of the client models.
The global update of the model at iteration \(t+1\) is given by
\begin{equation}
\label{eq:fedavg}
\texttt{Aggregate}(\theta_k^{(t)})=\texttt{FedAvg}(\theta_k^{(t)}):
\theta^{(t+1)} = \sum_{k=1}^{K} \frac{|\mathcal{D}_k|}{N} \theta_k^{(t)},
\end{equation}
where \( \theta_k^{(t)} \) is the model on the client \(k\), \( |\mathcal{D}_k| \) is the size of client \(k\)’s dataset, and \( N = \sum_{k=1}^{K} |\mathcal{D}_k| \) is the total data across all clients.
This technique is effective for general non-IID settings and reduces the need for frequent communication by allowing clients to perform multiple local updates before sending the model to the server~\cite{iyer2024review}.

A variation of the \textit{FedAvg}, \textit{FedMedian} is a robust aggregation method designed to handle adversarial or noisy updates.
Instead of computing the average of the model updates, the central server computes the coordinate-wise median of the client models.
If \( \theta_k^{(t)} \) is the local model update from client \(k\), then the global model update is
\begin{equation}
\texttt{Aggregate}(\theta_k^{(t)})=\texttt{FedMedian}(\theta_k^{(t)}):\theta^{(t+1)} = \text{median}(\theta_1^{(t)}, \theta_2^{(t)}, \dots, \theta_K^{(t)}).
\end{equation}
\textit{FedMedian} is particularly useful in scenarios where a subset of clients may contribute corrupted or malicious updates.
The impact of outliers is reduced using the median, resulting in more robust global models~\cite{yin2018byzantine}.

\textit{FedProx} (Federated Proximal), introduced by \citet{li2020federated}, extends \textit{FedAvg} to address the challenges of heterogeneity in FL systems.
\textit{FedProx} adds a proximal term to the local objective function to limit the deviation of local models from the global model.
The objective for client \(k\) is
\begin{equation}
\texttt{Aggregate}(\theta_k^{(t)})=\texttt{FedProx}(\theta_k^{(t)}):\min_{\theta_k} \left[ \mathcal{L}_k(\theta_k) + \frac{\mu}{2} \|\theta_k - \theta^{(t)}\|^2 \right],
\end{equation}
where \( \mu \) is a constant that controls the regularization strength and \(\mathcal{L}\) is the loss function.
\textit{FedProx} improves convergence in scenarios with non-IID data and clients with varying computational capabilities\cite{li2020federated}.

Another approach, \textit{FedOpt} (Federated Optimization), is a generalized optimization framework that builds on \textit{FedAvg} and introduces adaptive optimization techniques into FL.
\textit{FedAvg}'s update rule
can be rewritten to resemble
a stochastic gradient descent (SGD) update with learning parameter $\eta = 1$,
\begin{equation}
\theta^{(t+1)} = \theta^{(t)} + \sum_{k=1}^{K} \frac{|\mathcal{D}_k|}{N} ( \theta_k^{(t)} - \theta^{(t)}),
\end{equation}
where the deviations from the previous round's model, $g^{(t)}_k = ( \theta_k^{(t)} - \theta^{(t)} )$, are viewed as approximations of each client's gradient and the term $g^(t) = \sum_{k=1}^{K} \frac{|\mathcal{D}_k|}{N} g_k^{(t)}$ represents the average gradient across clients.

\textit{FedOpt} replaces the standard gradient descent in the expression above with other update rules such as momentum, AdaGrad, or Adam\cite{kingma2014adam}.
The server update rule is
\begin{equation}
\begin{aligned}
    \texttt{Aggregate}(\theta_k^{(t)})=\texttt{FedOpt}(\theta_k^{(t)}):\theta^{(t+1)} &= \theta^{(t)} + \eta v^{(t+1)}, \\  \; \text{where} \; v^{(t+1)} &= \mathcal{G}(g^{(t)}, \theta^{(t)}, v^{(t)}, \eta), \\
    g^{(t)} &= \sum_{k=1}^{K} \frac{|\mathcal{D}_k|}{N} (\theta_k^{(t)} - \theta^{(t)})
\end{aligned}
\end{equation}
where $\eta$ is a learning rate coefficient,
$v^{(t)}$ is called the momentum term
and $g^{(t)}$, as previously discussed, acts as an estimate of the average gradient.
The function $\mathcal{G}$ stands for a gradient-based update rule, such as AdaGrad or Adam.
Notice that by setting $\eta = 1$ and
$\mathcal{G}(g^{(t)}, \theta^{(t)}, v^{(t)}, \eta) = g^{(t)}$
we recover \textit{FedAvg}. 
\textit{FedOpt} improves the learning process in federated settings, mainly when there is a high variance in the client updates~\cite{reddi2020adaptive}.


Recent research in federated learning has made significant progress in tackling the challenges of long training times and high computational costs. By incorporating techniques like model compression, efficient aggregation, and asynchronous updates, researchers have found ways to make the process faster and more efficient. For instance, one approach adjusts local update frequencies and applies gradient compression to cut down on communication costs and speed up convergence.\cite{song2024joint} Another method, TEASQ-Fed, takes this further by using sparsification and quantization to reduce overhead in asynchronous settings.\citet{jia2024efficient} Similarly, frameworks like FedAT address the issue of the straggler effect by using asynchronous tiers and staleness-aware weighted aggregation to ensure more balanced contributions from different devices.\cite{chai2021fedat} BLADE builds on these ideas by incorporating adaptive pruning and quantization, which help minimize total training time.\cite{ying2024blade} These advancements show that by strategically combining compression techniques with asynchronous protocols, federated learning can become more scalable and efficient.
Beyond these methods, other aggregation techniques exist as well, including probabilistic models and security-focused approaches like blockchain-based solutions \cite{qi2024model}.

\subsection{Industrial Chemical Engineering Workflows with Federated Learning}

Federated learning (FL) has the potential to transform industrial chemical engineering by enabling advanced condition monitoring and fault diagnosis across geographically distributed plants.
Many industrial workflows rely on sensors to track parameters like temperature, pressure, and flow rates—data that is inherently non-IID due to differences in process designs, equipment configurations, and environmental conditions.
FL can allow each plant to train its own condition-based maintenance (CBM) models locally while sharing only encrypted model parameters for aggregation \cite{berghout2022federated}.
However, the deployment of FL in such settings is not without challenges.
Sensor calibration heterogeneity, sporadic fault occurrences, and variable data quality often lead to model drift and convergence instability.
Addressing these issues requires robust strategies such as adaptive aggregation techniques, local model personalization, and dynamic weighting schemes to mitigate the bias introduced by non-IID data.

At the same time, FL is driving progress in collaborative molecular property prediction, particularly in catalyst design and process optimization.
Research labs and industrial R\&D centers, each working with proprietary chemical datasets, leverage FL in combination with graph neural networks (GNNs) to model complex molecular interactions\cite{heyndrickx2023melloddy}.
Techniques like scaffold splitting and latent Dirichlet allocation (LDA) help simulate heterogeneous data environments \cite{zhu2022federated}.
Models such as FLIT(+)\cite{zhu2022federated}, which apply instance reweighting to account for sample uncertainty, demonstrate how distributed training can produce a robust global model that accurately predicts molecular properties while preserving data sovereignty.
Persisting challenges such as inconsistent molecular representations across datasets, imbalanced chemical data, and divergent local training objectives can lead to suboptimal global model performance.

Advanced FL architectures have been developed for industrial applications to tackle challenges in anomaly detection and large-scale model training.
One such method, TemporalFED \cite{gomez2023temporalfed}, integrates time-series conversion with feature engineering using autocorrelation functions and discrete Fourier transforms to extract high-order features from process data. This enables real-time detection of cyberattacks and process anomalies in chemical plants.
Another technique, Federated Opportunistic Block Dropout (FEDOBD), decomposes deep neural networks into semantic blocks and selectively transmits only critical updates based on the Mean Block Difference (MBD) metric \cite{chen2023efficient}.
This dropout mechanism reduces communication overhead in bandwidth-limited industrial networks, ensuring fast model convergence without sacrificing diagnostic accuracy. 

Despite these advances, real-world FL deployments in industrial chemical engineering still face hurdles.
Aligning FL with strict regulatory and data privacy requirements while integrating with legacy operational technology (OT) systems and diverse sensor networks requires careful planning and collaboration across disciplines \cite{berghout2022federated}. Moreover, industry-focused surveys such as \citet{manzoor2024survey} provide strategies that could be adapted to chemical engineering, where techniques such as secure multiparty computation (SMC) and differential privacy have been discussed to allow clients to perform local computations without revealing sensitive information. By adding carefully calibrated noise to model updates, differential privacy ensures that individual data contributions cannot be reverse-engineered from the aggregated results. This becomes an essential point for chemical engineering datasets, where each data point may reveal detailed operational insights.

Even with these challenges, FL offers a promising, privacy-preserving framework that enhances process safety and improves operational efficiency—paving the way for smarter, more resilient industrial chemical engineering workflows.

\subsection{Tools to implement Federated Learning Frameworks}

Several frameworks have been developed to facilitate the implementation of FL in practice. One such framework is \texttt{Flower}~\cite{beutel2020flower}, which provides a flexible environment to conduct FL experiments.
\texttt{Flower} supports multiple ML libraries and offers a modular design that allows users to customize their training workflows.

Another popular framework is \texttt{TensorFlow Federated (TFF)}~\cite{The_TensorFlow_Federated_Authors_TensorFlow_Federated_2018}, which is built on the widely-used TensorFlow\cite{tensorflow2015-whitepaper} library.
\texttt{TFF} allows developers to convert existing TensorFlow models into federated setups, making it accessible to those already familiar with TensorFlow’s API.
\texttt{TFF} is designed explicitly for scalable FL experiments, supporting both simulation and real-world distributed training scenarios. 

In this paper, we mainly focus on the \texttt{Flower} and \texttt{TensorFlow Federated} framework, exploring how they can be applied to chemical engineering tasks that require decentralized model training to preserve privacy.

\section{Tutorial on Federated Learning --- Pill classification for pharmaceutical manufacturing}

The purpose of this section is to put the theory discussed above into practice in a chemical engineering context.
Specifically, we build and train an FL model to detect, based on visual information, whether produced pills are acceptable or defective.
We present a data set of relevance, followed by a description of the implementation of an FL framework in both \texttt{Flower} and \texttt{TensorFlow Federated}, and finalize with the numerical results obtained after executing these codes.

Quality assurance is of utmost importance for pharmaceutical manufacturing ~\cite{ICH_Q10_Guideline}.
For example, ensuring the detection of anomalous pills during production is necessary to safeguard consumer safety and comply with government regulations~\cite{FDA_cGMP_21st_century, FDA_CFR_210_1}.
However, manufacturers may be reluctant to share data on potential defects due to security concerns.
In this scenario, FL emerges as a solution for secure collaboration.
Manufacturers act as clients, using their data to train models locally on how to detect pill anomalies
while sharing only the resulting models with a central server.

\begin{figure}[tbh]
    \centering
    \subfloat[Acceptable pills.]{
        \label{fig:pills_good}
        \includegraphics[width=0.2\textwidth]{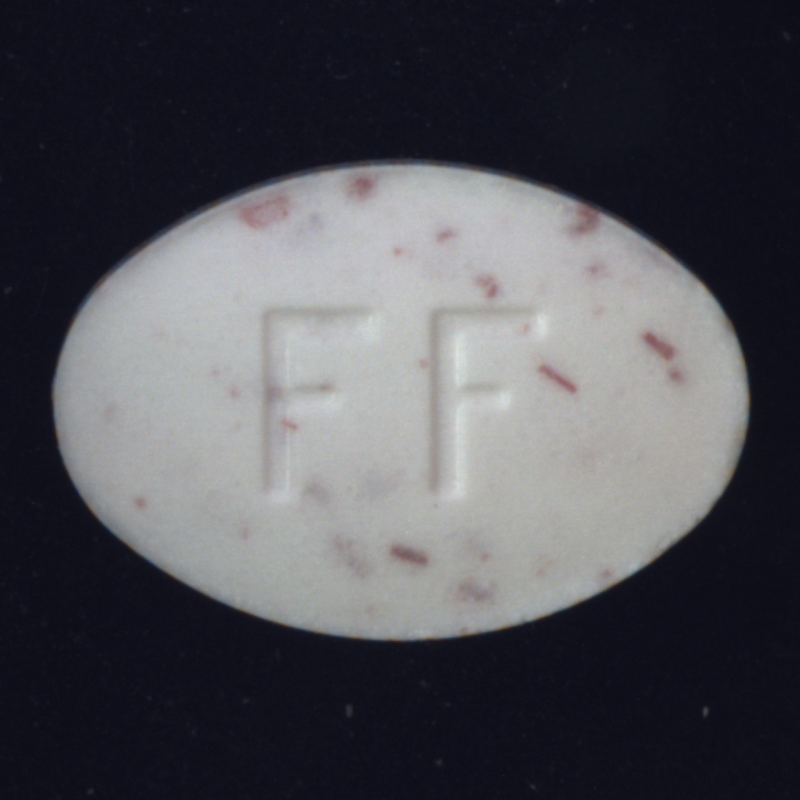}
        \qquad
        \includegraphics[width=0.2\textwidth]{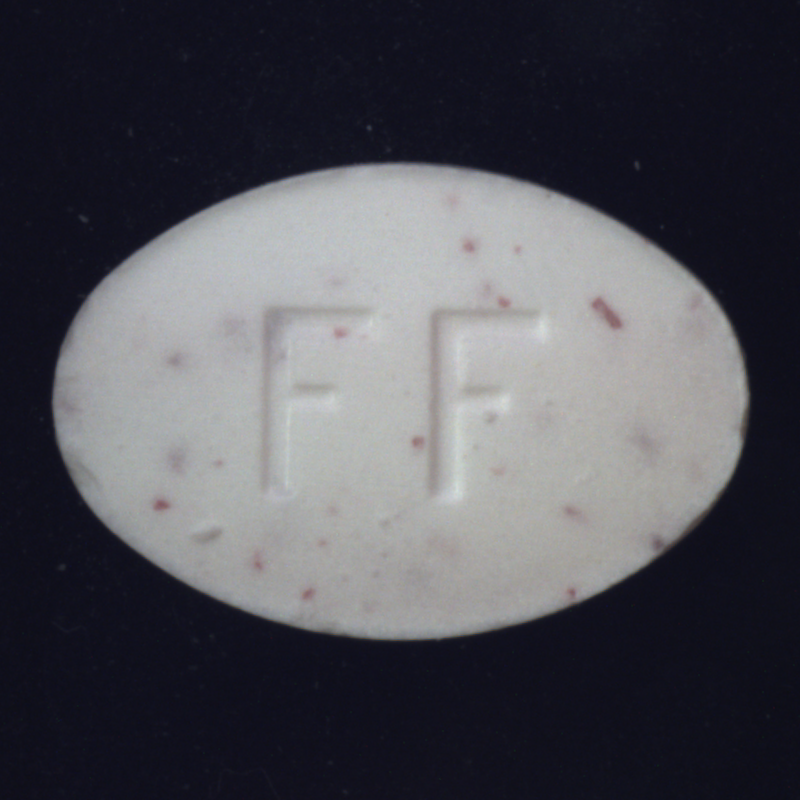}
        \qquad
        \includegraphics[width=0.2\textwidth]{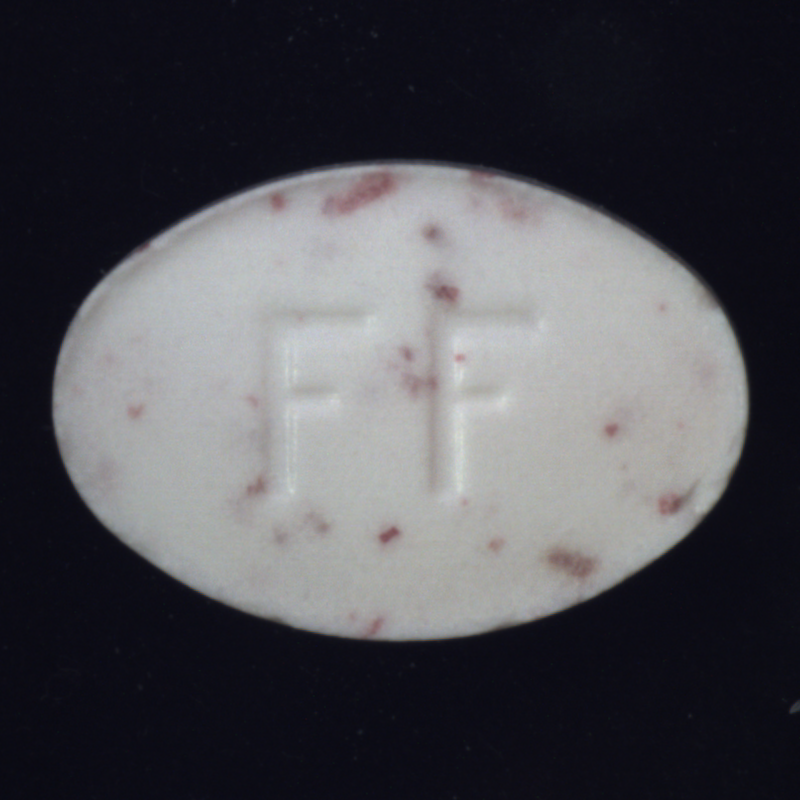}
    } 
    
    \subfloat[Defective pills.]{
        \label{fig:pills_bad}
        \includegraphics[width=0.2\textwidth]{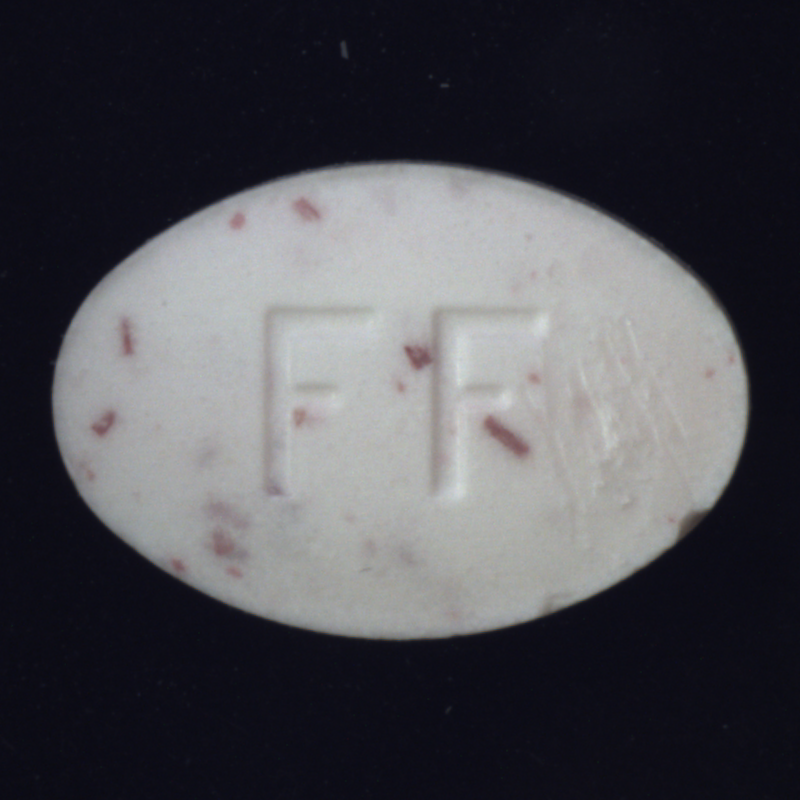}
        \qquad
        \includegraphics[width=0.2\textwidth]{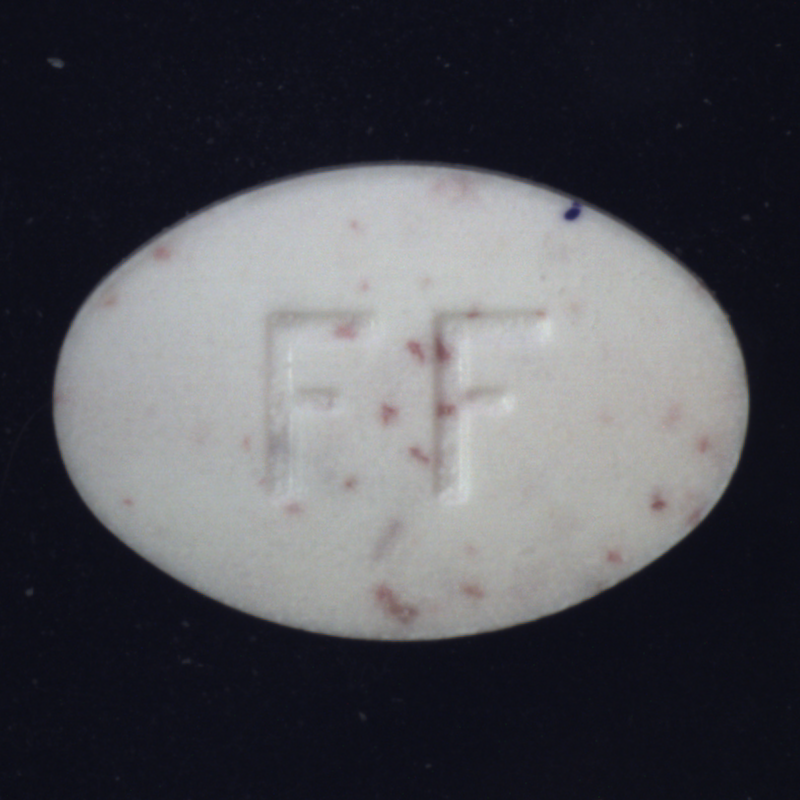}
        \qquad
        \includegraphics[width=0.2\textwidth]{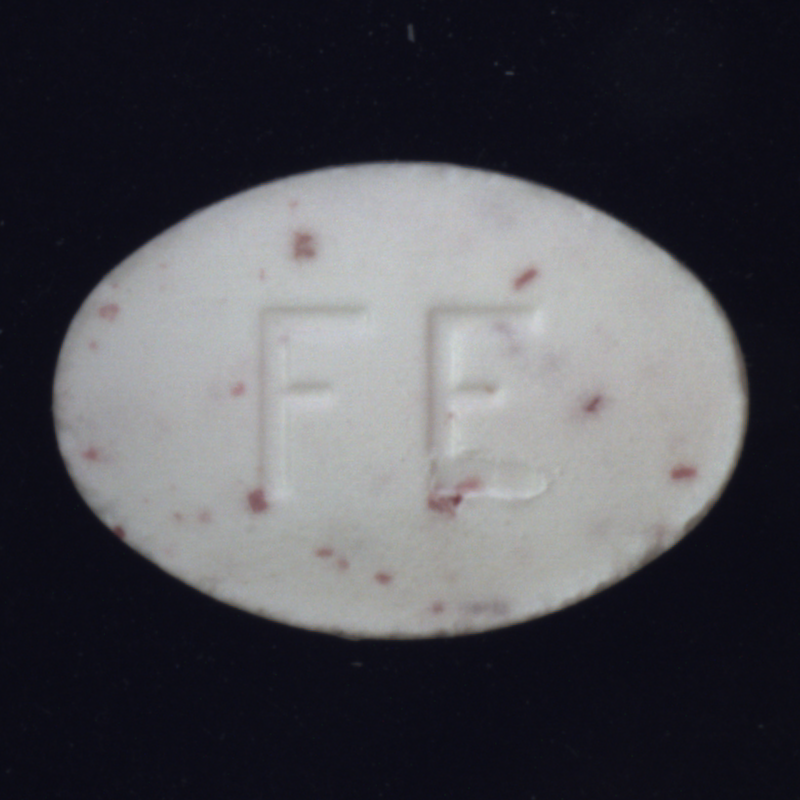}
    }
\caption{Example pills taken from the dataset~\cite{Bergmann_2019, Bergmann_2021}. The pills above are defective due to scratches, color changes, and faulty imprints, respectively. Notice that for a lay observer, it may be difficult to distinguish between faulty and good pills.}
\label{fig:pills}
\end{figure}

We use images from the MVTec Anomaly Detection (MVTec AD) dataset~\cite{Bergmann_2019, Bergmann_2021}.
The dataset comprises 434 photos of the same pill variety, consistently aligned and photographed against a uniform dark background.
The data is separated into two classes:
one for acceptable pills and the other for pills with a range of possible defects, such as color changes, contamination, cracks, faulty imprints, and scratches.
See Fig.~\ref{fig:pills} for examples of images from the dataset.

This is a binary classification task,
and the training process will employ the classifier's \emph{cross-entropy}~\cite{Good_1952} as the loss function for all clients, a standard practice in classification Machine Learning models.
The server weight aggregation is done through the \textit{FedAvg} algorithm, explained in Equation~\eqref{eq:fedavg}.

\subsection{Federated Learning Implementation using Flower framework}

The FL pipeline implemented using \texttt{Flower}\cite{beutel2020flower} consists of several essential components, each designed to allow multiple distributed clients to collaboratively train a global ML model without sharing their local data.
The process begins with the definition of a central server and a strategy to manage communication between clients and the server.
A custom strategy, \texttt{FedCustom}, derived from \texttt{Flower}'s base \texttt{Strategy}\cite{mcmahan2023communicationefficientlearningdeepnetworks} class, is created to handle client sampling, parameter aggregation, and client model evaluation.

The first step is to set up a client function that encapsulates the model architecture, optimizer configuration, and client-side operations.
Each client is represented by the \texttt{FlowerClient} class, which defines how a client interacts with the global server.
This class includes three primary methods: \texttt{get\_parameters}, \texttt{fit}, and \texttt{evaluate}.
These methods manage the download of model weights from the server, the local training on client-specific data, and the upload of updated weights to the server after each training round.
The \texttt{FlowerClient} is constructed from arguments such as client ID, model architecture, and data loaders for training and validation.
The client participates in the FL process by training its local model and sending the updates to the server.

The server employs the \texttt{FedCustom} strategy, detailed below, to coordinate training rounds, sample clients, and aggregate model updates.
The global model parameters are initialized with each training round configured.
This configuration includes sampling a fraction of clients (\texttt{fraction\_fit}) and setting training hyperparameters, such as the learning rate and the number of local epochs.
At the end of each round, client model updates are aggregated using a weighted average of parameters.
The aggregation ensures that clients contributing more data have a proportionally higher influence on the global model update.
As mentioned above, the \textit{FedAvg} parameter aggregation procedure on the server is the most commonly used, yet the framework allows for the implementation of custom aggregation rules.


The federated system must handle variability in client data distribution, computational capacity, and network conditions.
The \texttt{FedCustom} strategy manages such variability, including client failures, by setting \texttt{accept\_failures=True}\cite{beutel2020flower}.
This setting ensures that training continues even if some clients fail to participate in specific rounds.
The strategy also allows dynamic configuration of local hyperparameters, which adjusts training parameters such as the learning rate and the number of epochs based on the current round.
Evaluation is handled similarly, allowing clients to locally test their models on validation data and report metrics such as loss and accuracy.
Algorithm \ref{alg:2} exemplifies the procedure for the client-side training process.

\begin{algorithm}
\caption{Client Training Process}
\label{alg:2}
\begin{algorithmic}[1]
\STATE Initialize local model with global parameters $\theta_{\text{global}}$
\STATE Configure optimizer with learning rate $\eta$ and loss function $\mathcal{L}$
\FOR{each epoch $e = 1, 2, \dots, E$}
    \FOR{each batch of local data}
        \STATE Perform forward pass: $\hat{y} = f(x; \theta)$
        \STATE Compute loss: $\mathcal{L}(\hat{y}, y)$
        \STATE Compute gradients: $\Delta \theta = \frac{\partial \mathcal{L}}{\partial \theta}$
        \STATE Update model weights: $\theta = \theta - \eta \Delta \theta$
    \ENDFOR
\ENDFOR
\STATE Return updated parameters $w$ to the server
\end{algorithmic}
\end{algorithm}

In distributed settings,
the clients must train the same models
but can choose different hyperparameters,
to better suit their data quality or computational resources.
To represent this,
we let the learning rates be client-specific.

Finally, after setting up the global model, data loaders, and the federated strategy, the simulation is initiated using \texttt{Flower.simulation.start\_simulation}\cite{beutel2020flower}.
This command triggers federated training, in which each client independently trains its model and periodically synchronizes with the global server.
The server aggregates client models across several training rounds, progressively refining the global model.
FL ensures privacy in this context by not sharing raw data with clients and complying with strict privacy regulations common in chemical engineering and related fields.

\subsection{Federated Learning Implementation with TensorFlow Federated}

The FL pipeline implemented using \texttt{TensorFlow Federated (TFF)}\cite{The_TensorFlow_Federated_Authors_TensorFlow_Federated_2018} enables multiple distributed clients to collaboratively train a global ML model on decentralized data, ensuring privacy and data security.
In this implementation, we simulate an FL system with multiple clients using image data to classify pill defects.
The system is designed with core components such as the client model, the federated training process, and the evaluation strategy.

To begin with, the data set is loaded and partitioned between multiple clients.
Each client receives a portion of the dataset for local training, simulating the decentralized data scenario in FL.
The core ML model for pill classification is built using TensorFlow's \texttt{Sequential} API\cite{tensorflow2015-whitepaper}.
This model, named \texttt{PillModel}, consists of convolutional layers for feature extraction followed by fully connected layers for classification.
The model concludes with a dense layer and a softmax activation to classify the pill images into two categories.
The input specification, \texttt{input\_spec}, defines the expected input shape and data types, ensuring compatibility with the FL framework of \texttt{TFF}.

The FL process is managed by \texttt{TFF}'s \texttt{build\_weighted\_fed\_avg} function, which implements \textit{FedAvg}\cite{mcmahan2023communicationefficientlearningdeepnetworks} algorithm.
The training process begins with initializing the global model and performing several rounds of communication between the server and the clients.
In each round, a subset of clients (in this case, all clients) is selected to participate.
Each client trains the model on its local dataset using the Adam\cite{kingma2014adam} optimizer, and the updated model weights are sent back to the central server.
The server aggregates these updates using weighted averaging based on the number of samples each client has. The aggregation algorithm is the same as present in the \texttt{Flower} framework.

Client-side training is handled locally by each client on its partitioned data.
After each training round, the model is evaluated on a separate test dataset.
The evaluation is done using the \texttt{federated\_evaluate} function, which measures the model's performance on unseen data.
The evaluation process initializes a fresh evaluation model and copies the trained global weights from the federated training process.
Each client evaluates the model on its local test data, and the results are aggregated to provide a centralized evaluation metric.
This procedure ensures that the model performance is tested without clients sharing their data as described in Algorithm~\ref{alg:2}, implemented similarly as in the \texttt{Flower} framework.

\subsection{Computational results}

For this example and the upcoming ones, we implement both a federated learning approach and a centralized approach. The centralized method is trained over 200 training epochs, whereas the federated learning approach is trained for 20 communication rounds, each with 10 epochs per client with 10 clients each round.
This guarantees that both methods will visit the data the same number of times.
Although we also discussed a \texttt{TFF} implementation for completeness,
all showcased results were obtained using the \texttt{Flower} framework.

In addition, we use confusion matrices and receiver operating characteristic (ROC) curves to report our results.
The confusion matrices provide a detailed summary of the model's classification performance by breaking down the predictions into four categories: True Positives (TP), False Positives (FP), True Negatives (TN), and False Negatives (FN).
The matrix's entries correspond to the number of predictions falling into these categories, allowing us to evaluate the model's ability to distinguish between classes.

The confusion matrices serve as a diagnostic tool for the presented results, identifying where the model performs well and where it struggles.
High values along the diagonal indicate strong classification performance, whereas off-diagonal values reveal instances of misclassification.
The distribution of values in the confusion matrices can also provide insights into class imbalance or other biases present in the dataset, which could inform subsequent modifications to the training process or model architecture.

ROC curves evaluate the trade-off between the True Positive Rate (TPR) and the False Positive Rate (FPR) at various classification thresholds. Each curve visually depicts how a classifier's performance changes as the decision boundary is adjusted.

The results include multiple types of ROC curves:
\begin{itemize}
    \item Individual class ROC curves as explained above.
    \item Worst-Case Line: This line represents a baseline random classifier that performs no better than random guessing. It serves as a reference point, indicating that any ROC curve above this line signifies a classifier that performs better than random.
    \item Micro-Average ROC Curve: The micro-average ROC curve aggregates the contributions of all classes and computes the average ROC curve by considering each instance equally, regardless of the class it belongs to. This method is proper when we want to evaluate the overall performance of the model across all classes, providing a single ROC curve that represents the collective classification capability.
    \item Macro-Average ROC Curve: The macro-average ROC curve, on the other hand, computes the ROC curve for each class individually and then averages these curves. Unlike the micro-average, the macro-average gives equal weight to each class, making it more sensitive to the performance of minority classes. This can be beneficial for assessing the model's ability to handle class imbalance.
\end{itemize}

Each ROC curve is accompanied by its corresponding area under the curve (AUC), a scalar value that summarizes the model's performance.
Higher AUC values indicate better classification capability, with 1.0 representing perfect classification and 0.5 corresponding to random guessing.
The micro and macro AUC values provide complementary perspectives: micro-AUC emphasizes overall performance, while macro-AUC highlights balanced performance across classes.

The confusion matrices show that the model performs well for most classes but reveal specific areas for improvement, particularly in reducing False Positives for specific categories.
The ROC curves demonstrate that the model consistently outperforms the worst-case baseline, indicating robust performance.
The separation between the micro-average and macro-average ROC curves suggests that the model may perform better for the majority classes, with room for optimization in handling minority classes more effectively.
These results collectively provide a comprehensive evaluation of the model's classification performance, allowing for targeted enhancements based on the observed patterns.

\begin{figure}[tbh]
    \centering
    \subfloat[Centralized]{
        \includegraphics[width=0.48\textwidth]{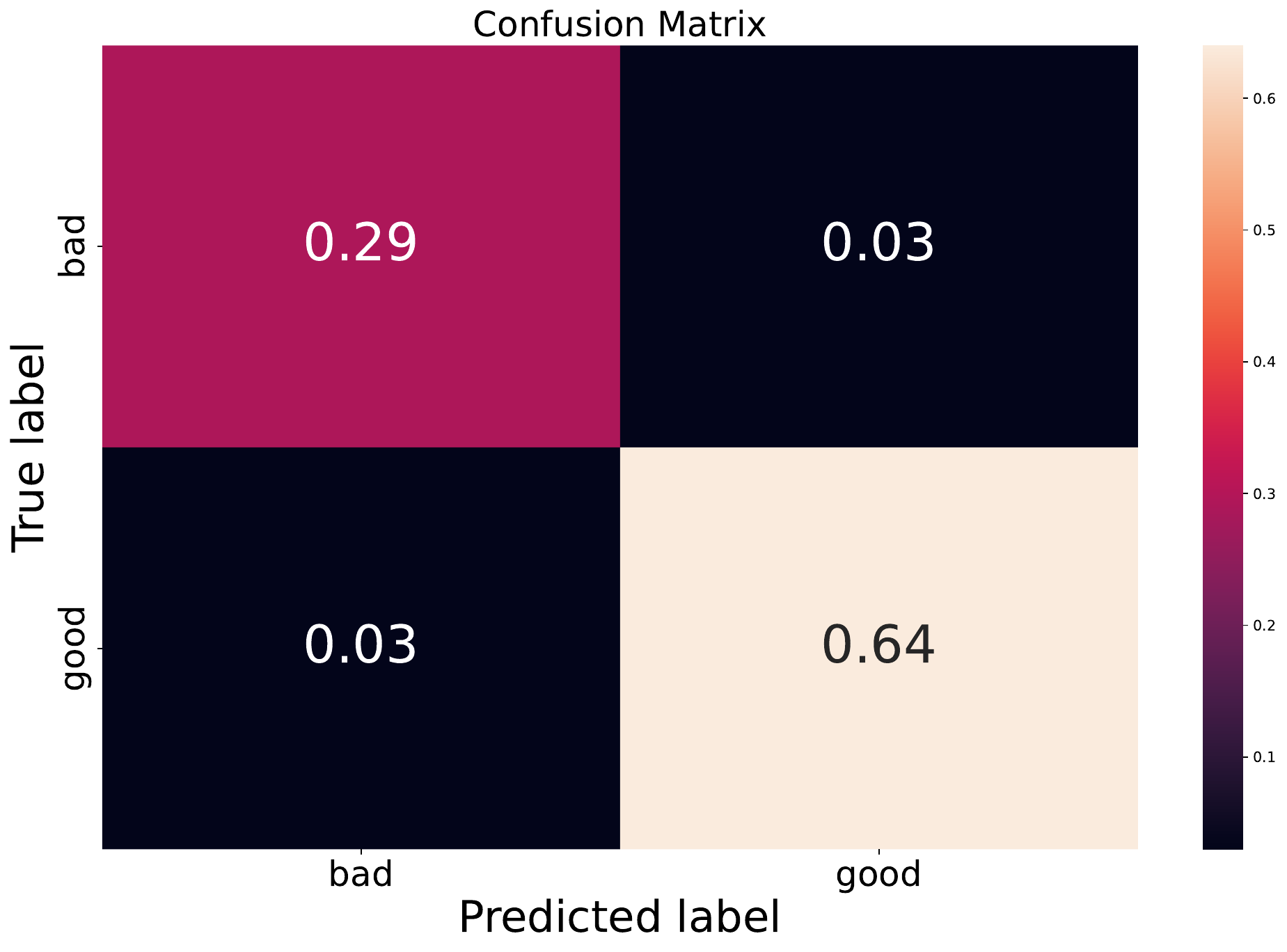}
    } 
    \hfill
    \subfloat[Federated]{
        \includegraphics[width=0.48\textwidth]{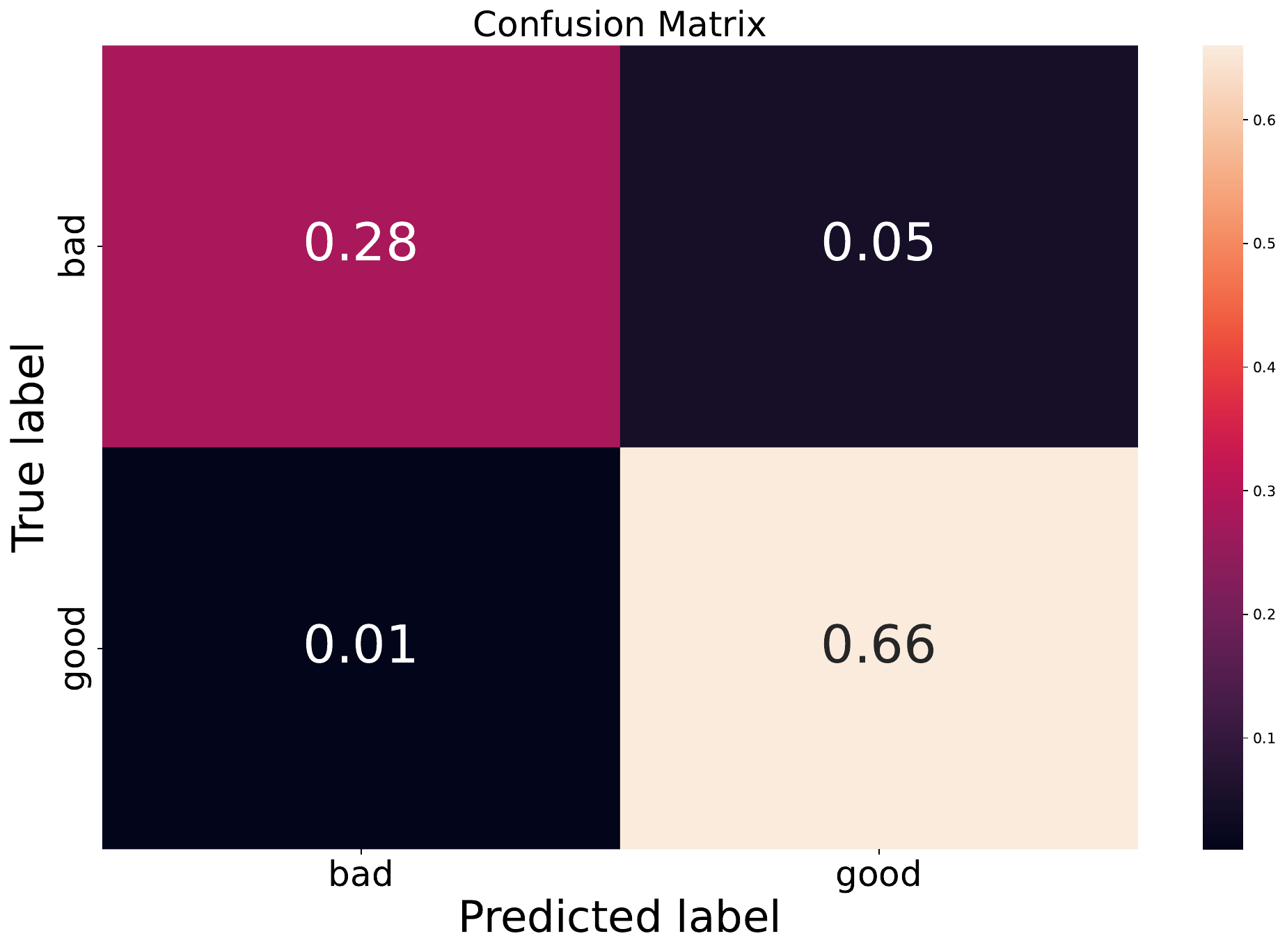}
    }
    \caption{Confusion matrix for the Pill dataset trained with \texttt{Flower}.}
    \label{fig:pill_confusion_matrix}
\end{figure}

\begin{figure}[tbh]
    \centering
    \subfloat[Centralized]{
        \includegraphics[width=0.48\textwidth]{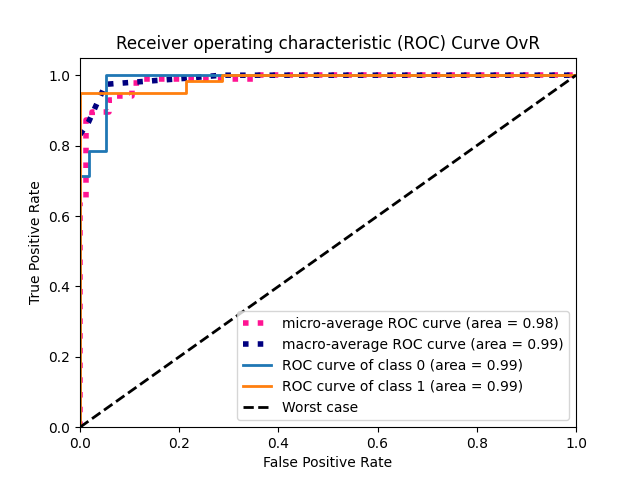}
    } 
    \hfill
    \subfloat[Federated]{
        \includegraphics[width=0.48\textwidth]{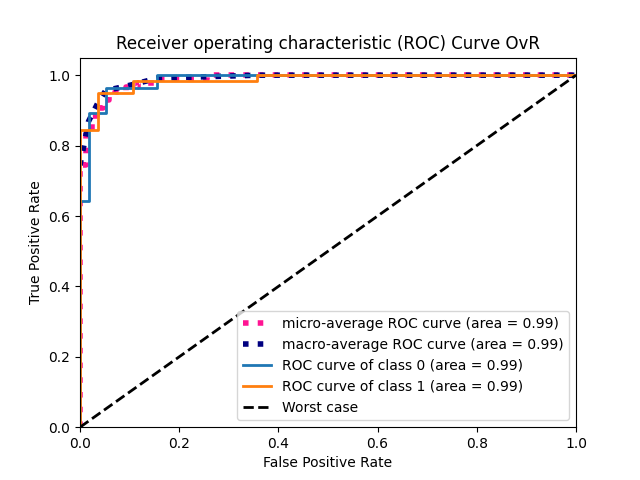}
    }
    \caption{Receiver operating characteristic curve (ROC) for the Pill dataset.}
    \label{fig:pill_roc}
\end{figure}

For the pill dataset, we have the comparison between the centralized and federated approach using confusion matrices in Figs.~\ref{fig:pill_confusion_matrix} and \ref{fig:pill_roc}.
Comparing the two confusion matrices, the federated approach demonstrates an improvement over the centralized approach.
The federated model gives similar results to the centralized one, with slight variation in the classification accuracy for "good" (66\% vs. 64\%) and "bad" (28\% vs. 29\%) labels and the same total misclassification rates.
In particular, the federated model only misclassifies 1\% of "good" labels as "bad" compared to 3\% in the centralized model, while it misclassifies 5\% of "bad" labels as "good" versus 3\% in the centralized case.
These results indicate that, given the same number of passes over the data, the federated approach can capture the nuances of each class even without direct client communication.
This results in accurate decision boundaries and a slight increase in true positives and negatives. In general, the federated approach provides a robust classification model, especially for distinguishing between minority classes.

The ROC curves for the federated learning approach also demonstrate that it keeps up with the centralized approach.
In the federated model, both the micro-average and macro-average ROC curves achieve an area of 0.99, indicating near-perfect classification performance for both classes, compared to 0.98 (micro) and 0.99 (macro) in the centralized model.
Additionally, the class-specific ROC curves for the federated model have areas of 0.99, the same observed in the centralized case.
The near-perfect alignment of the federated ROC curves with the top-left corner indicates a high true positive rate and minimal false positives, highlighting that the federated model is effective at distinguishing between classes.
This suggests that federated learning can take advantage of the diversity of data across distributed sources, leading to a robust and accurate classifier. This aligns with the findings of comparative studies\cite{majeed2022comparative,lazaros2024federated}, which empirically demonstrate that FL’s decentralized training on non-IID data, common in real-world scenarios, leads to stronger generalization compared to centralized methods. Centralized models often struggle with data homogeneity and fail to adapt to local variations, resulting in suboptimal decision boundaries. For instance, a systematic review highlights that FL’s decentralized training on non-IID data enhances model generalization, especially in NLP tasks like sentiment analysis and text classification\cite{khan2024federated}.

In general, the comparison of the confusion matrices and the ROC curves reveals that the federated learning approach performs equivalently to the centralized model in classification accuracy and robustness.
The federated model not only reduces misclassification errors, as shown by the confusion matrices but also achieves near-perfect ROC areas of 0.99 for both micro and macro averages, indicating high-class separation and consistent performance across both classes.

The results corresponding to this tutorial, which we call the PILL dataset, show that in the centralized experiments, the model achieved a training accuracy of 99.69\% and a test accuracy of 93.27\% with a test loss of 0.1576.
In the federated experiments, the model obtained a lower training accuracy of 93.54\% but showed an improvement in the test accuracy, reaching 94.79\%, and a test loss of 0.207.
Although federated learning demonstrated better generalization and stability for the test set, the training time also decreased from 3019.34 seconds in the centralized case to 2215.40 seconds in the federated case, indicating that federated learning can enhance performance for the same number of passes over the data.
These results suggest that federated learning offers equivalent classification performance and robustness for the PILL dataset, making it a suitable approach for applications that require data-privacy over distributed data.
These results and other examples relevant to chemical engineering are presented in Table~\ref{tab:experiment-results}.


\section{Results for Federated Learning relevant to Chemical Engineering}

In this section, we introduce other case studies in which FL is applied to cases relevant to chemical engineering.
In particular, we are going to discuss two examples.
The first is a multimodal mixture of experts, where genomic and image data are used together to train a biological classifier. 
The second example is an application in drug discovery, where graph neural networks (GNNs) in local clients are trained to find promising chemical substances for HIV treatments.

As for the PILL dataset,
these are multiclass classification tasks.
Unless otherwise stated, the loss function for all clients is cross-entropy, and
the server weight aggregation is done through the \textit{FedAvg} algorithm, previously explained in Equation~\eqref{eq:fedavg}.

\subsection{Multimodal Mixture of Experts}

\begin{figure}[tbh]
    \centering
    \begin{adjustbox}{width=\textwidth}
\begin{tikzpicture}
[
    every node/.style = {align = flush center},
    every new ->/.style = {-Latex},
    box/.style = {
        fill = orange!20,
        inner sep = 4mm,
        rectangle,
        rounded corners,
        align = flush center,
    },
    input/.style = {
        box,
        fill = green!40,
    },
    output/.style = {
        box,
        fill = red!40,
    },
    conv/.style = {minimum width = 1.5cm},
    maxpool/.style = {
        minimum width = 1.2cm
    },
    para/.style = {
        parallelepiped,
        minimum height=2.3cm,
        minimum width=0.7cm, 
        thick,
    },
    fcl/.style = {
        para,
        fill = green!40,
        draw = green!80!black,
    },
    linear/.style = {
        para,
        fill = blue!40,
        draw = blue!60!black,
    },
    conv layer/.pic = {
        \foreach [count=\i] \X/\Col in {-0.2/red, 0/green, 0.2/blue} {
            \draw[canvas is yz plane at x = \X, transform shape, draw = black, fill = \Col!50!white, opacity = 0.5] 
                (-1,0.5) rectangle (1,-0.5);
        }
    },
    maxpool layer/.pic = {
        \draw[canvas is yz plane at x = 0, transform shape, draw = black, fill = orange!50!white, opacity = 0.2] 
            (-1,-0.5) rectangle (1,0.5);
        \draw[canvas is yz plane at x = 0, transform shape, draw = black]
            (-1,-0.5) rectangle (0, 0)
            (1,-0.5)  rectangle (0, 0)
            (-1,0.5)  rectangle (0, 0)
            (1,0.5)   rectangle (0, 0);
    },
]

\makeatletter
\pgfkeys{/pgf/.cd,
  parallelepiped offset x/.initial=2mm,
  parallelepiped offset y/.initial=2mm
}
\pgfdeclareshape{parallelepiped}
{
  \inheritsavedanchors[from=rectangle] 
  \inheritanchorborder[from=rectangle]
  \inheritanchor[from=rectangle]{north}
  \inheritanchor[from=rectangle]{north west}
  \inheritanchor[from=rectangle]{north east}
  \inheritanchor[from=rectangle]{center}
  \inheritanchor[from=rectangle]{west}
  \inheritanchor[from=rectangle]{east}
  \inheritanchor[from=rectangle]{mid}
  \inheritanchor[from=rectangle]{mid west}
  \inheritanchor[from=rectangle]{mid east}
  \inheritanchor[from=rectangle]{base}
  \inheritanchor[from=rectangle]{base west}
  \inheritanchor[from=rectangle]{base east}
  \inheritanchor[from=rectangle]{south}
  \inheritanchor[from=rectangle]{south west}
  \inheritanchor[from=rectangle]{south east}
  \backgroundpath{
    \southwest \pgf@xa=\pgf@x \pgf@ya=\pgf@y
    \northeast \pgf@xb=\pgf@x \pgf@yb=\pgf@y
    \pgfmathsetlength\pgfutil@tempdima{\pgfkeysvalueof{/pgf/parallelepiped offset x}}
    \pgfmathsetlength\pgfutil@tempdimb{\pgfkeysvalueof{/pgf/parallelepiped offset y}}
    \def\ppd@offset{\pgfpoint{\pgfutil@tempdima}{\pgfutil@tempdimb}}
    \pgfpathmoveto{\pgfqpoint{\pgf@xa}{\pgf@ya}}
    \pgfpathlineto{\pgfqpoint{\pgf@xb}{\pgf@ya}}
    \pgfpathlineto{\pgfqpoint{\pgf@xb}{\pgf@yb}}
    \pgfpathlineto{\pgfqpoint{\pgf@xa}{\pgf@yb}}
    \pgfpathclose
    \pgfpathmoveto{\pgfqpoint{\pgf@xb}{\pgf@ya}}
    \pgfpathlineto{\pgfpointadd{\pgfpoint{\pgf@xb}{\pgf@ya}}{\ppd@offset}}
    \pgfpathlineto{\pgfpointadd{\pgfpoint{\pgf@xb}{\pgf@yb}}{\ppd@offset}}
    \pgfpathlineto{\pgfpointadd{\pgfpoint{\pgf@xa}{\pgf@yb}}{\ppd@offset}}
    \pgfpathlineto{\pgfqpoint{\pgf@xa}{\pgf@yb}}
    \pgfpathmoveto{\pgfqpoint{\pgf@xb}{\pgf@yb}}
    \pgfpathlineto{\pgfpointadd{\pgfpoint{\pgf@xb}{\pgf@yb}}{\ppd@offset}}
  }
}
\makeatother

\matrix[row sep=1cm, column sep={8mm, between borders}] {
    \node (mri-in)      [input]   {MRI\\Input}; & 
    \pic (mri-conv-pic) {conv layer};
    \node (mri-conv)    [conv]    {}; &
    \pic (mri-maxpool-pic) {maxpool layer};
    \node (mri-maxpool) [maxpool] {}; &
    \node (mri-lin)     [linear]  {}; &
    & & & &[6mm] 
    \node (mri-out) [output] {MRI\\Output}; & 
    \\
    & & & &
    \node (mix-concat) [box] {Feature\\Concatenation}; &
    \node (mix-mha)    [box] {Multi-Head\\Attention}; &
    \node (mix-full)   [fcl] {}; &
    \node (mix-wsum)   [box] {Weighted\\Sum}; &
    \\
    \node (dna-in)   [input]  {DNA\\Input}; &
    \node (dna-lin1) [linear] {}; &
    \node (dna-lin2) [linear] {}; &
    \node (dna-full) [fcl]    {}; &
    & & & & 
    \node (dna-out) [output] {DNA\\Output}; & 
    \\
};


\node[below = 1.2cm of mri-conv] {Conv2d\\+\\ReLU};
\node[below = 1.2cm of mri-maxpool] {MaxPool};
\node[below = 1mm of mri-lin] {2 Linear\\+\\ReLU};

\node[below = 1mm of dna-lin1] {2 Linear\\+\\ReLU};
\node[below = 1mm of dna-lin2] {2 Linear\\+\\ReLU};
\node[below = 1mm of dna-full] {Fully\\Connected\\+\\ReLU};
\node[below = 1mm of mix-mha] {(11 heads)};
\node[below = 1mm of mix-full] {Fully\\Connected\\+\\SoftMax};

\node[inner sep = 0.1mm, circle, fill=black, right = 6mm of mix-wsum] (fork) {};

\graph[use existing nodes,
       simple,
       every new ->/.style = {
        shorten >=1pt,
        -Latex, 
       },
       edges = {
        thick,
       },
       hv/.style = {to path={-| (\tikztotarget)}},
       vh/.style = {to path={|- (\tikztotarget)}},
] {
{ (mri-in) -> (mri-conv) -> (mri-maxpool) -> (mri-lin)
, (dna-in) -> (dna-lin1) -> (dna-lin2) -> (dna-full)
}
->[hv] (mix-concat) -> (mix-mha) -> (mix-full) -> (mix-wsum);
(mix-wsum) -- (fork) ->[vh] {(mri-out), (dna-out)};
{(mri-lin), (dna-full)} ->[hv] (mix-wsum);
};
\end{tikzpicture}
\end{adjustbox}
    \caption{The architecture for the DNA+MRI MMoE network starts by feeding the input to separate specialized networks, followed by both a gating network and a combination of their weights,
    and, finally, a shared layer before splitting the outputs into different modalities.
    Notice that while both inputs and outputs are separate, the hidden layers are shared between DNA and MRI.}
\label{fig:mmoe}
\end{figure}

In this subsection, we focus on the integration of DNA sequence data~\cite{singh2023b} and magnetic resonance imaging (MRI) scans~\cite{msoud_nickparvar_2021} into a Multimodal Mixture of Experts (MMoE) learning framework~\cite{yu2023mmoe}. Initially, we trained a model solely on DNA sequence data, with results discussed in the following sections. However, to enhance the model's versatility, we now aim to incorporate MRI data for brain tumor classification. This introduces the challenge of effectively integrating this new modality into the existing framework. The primary motivation for combining these two modalities within a single neural network is to build an expert biological model capable of handling a broad range of tasks. By integrating both genomics and high-dimensional imaging data, we propose to create a unified model that can process and understand diverse biological signals. This would enable the model to perform tasks such as the classification of brain tumors from MRI scans alongside the prediction of gene families from DNA sequences, effectively addressing the challenge.
This MMoE network approach allows the model to learn cross-modal relationships, enhancing representation power by capturing complementary features from the distinct domains of genomics and medical imaging~\cite{chan2022combining}.
In Fig.~\ref{fig:mmoe}, we show a high-level description of the architecture used in this MMoE network.

The MRI data set includes four classes of brain tumors: glioma, meningioma, non-tumor, and pituitary.
The model is tasked with accurately classifying these tumor types, highlighting its capability to handle large and complex medical imaging data.
In addition, the DNA data set consists of seven classes of gene families, each associated with distinct biological functions.
Class 0 includes G protein-coupled receptors (GPCRs), key players in cellular signaling processes.
Classes 1 and 2 represent protein kinases and phosphatases, enzymes responsible for phosphorylation and dephosphorylation, respectively.
Classes 3 and 4 correspond to enzymes that synthesize and catalyze the synthesis of various biomolecules.
Class 5 covers ion channels, essential for cell electrical signaling, while Class 6 encompasses transcription factors that regulate gene expression.
Before integrating the DNA sequences into the learning framework, we employ the TfidfVectorizer\cite{scikit-learn} with an n-gram range set to (5, 5). This configuration focuses on capturing pentameric (5-mer) patterns within the sequences, transforming them into a numerical format well-suited for neural network input. Using the TF-IDF (Term Frequency-Inverse Document Frequency) method, we effectively highlight the frequency and significance of these 5-mers, effectively encoding the nucleotide sequences in a format suitable for neural network input.

The MMoE architecture is designed with two branches: one for processing MRI images and the other for handling DNA vectors.
The MRI branch employs convolutional neural networks (CNNs) to extract spatial features from medical images, while the DNA branch uses fully connected layers to process the vectors generated from DNA sequences.
Later, these branches are concatenated, and a gating layer is applied, allowing the model to learn from both types of data jointly.
We present this architecture for local clients in Fig.~\ref{fig:mmoe}.
Our results suggest that the cross-modal interactions between DNA and MRI features help build a comprehensive biological representation, facilitating improved predictive performance for tasks that span both imaging and genetic data.

Using FL via the \texttt{Flower} framework, this MMoE model is trained in a distributed manner.
Each client trains the model locally to provide both the MRI and DNA data without sharing raw data, ensuring privacy and HIPPA compliance~\cite{topaloglu2021pursuit} with the regulations of medical data.
The server coordinates the aggregation of the client models, combining them into a single global model that progressively improves over multiple rounds of training.
This process preserves the privacy of sensitive medical and genetic information while leveraging the full potential of multimodal learning for healthcare applications.

\begin{figure}[tbh]
    \centering
    \subfloat[Centralized DNA]{
        \includegraphics[width=0.48\textwidth]{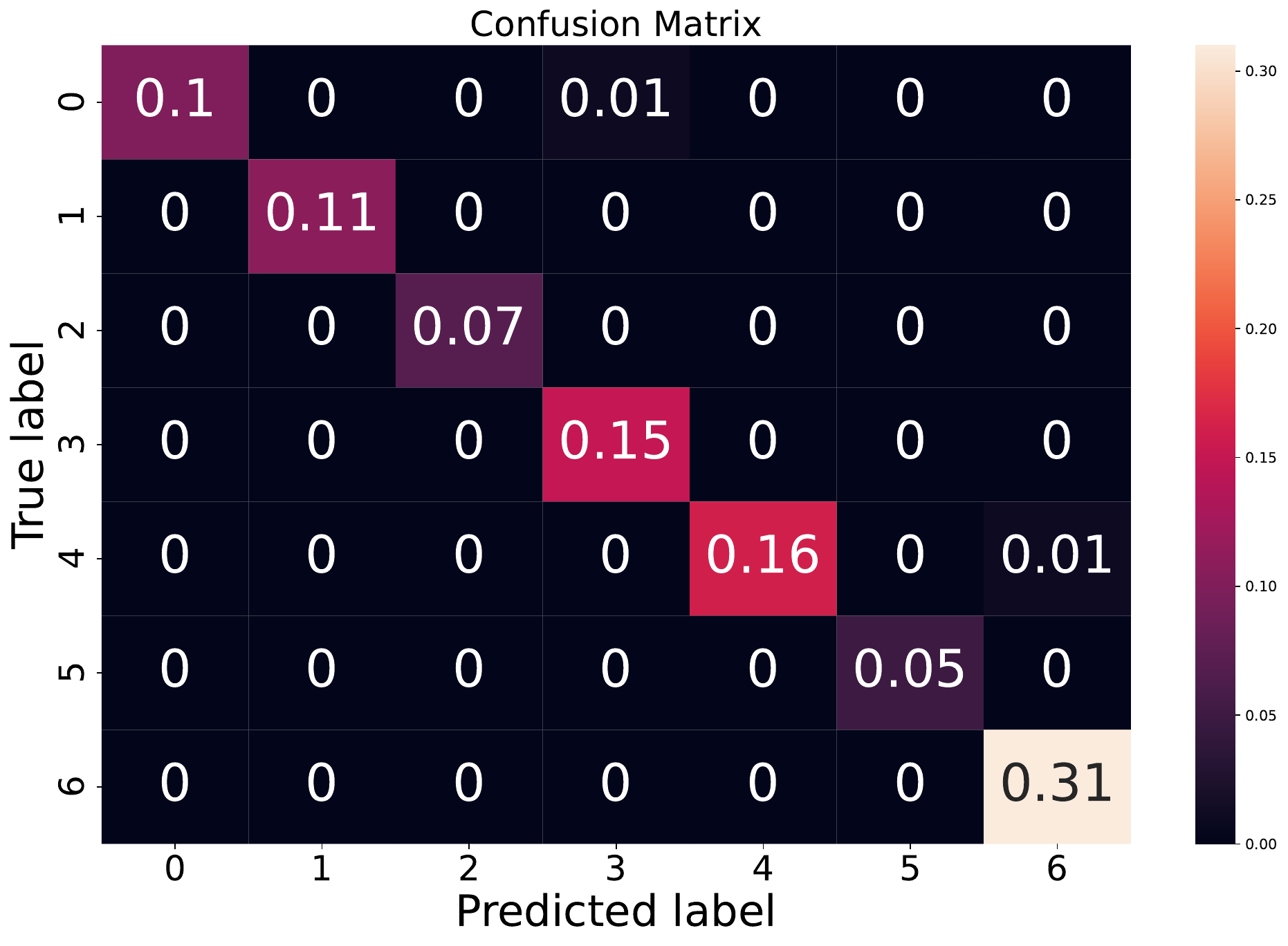}
    } 
    \hfill
    \subfloat[Federated DNA]{
        \includegraphics[width=0.48\textwidth]{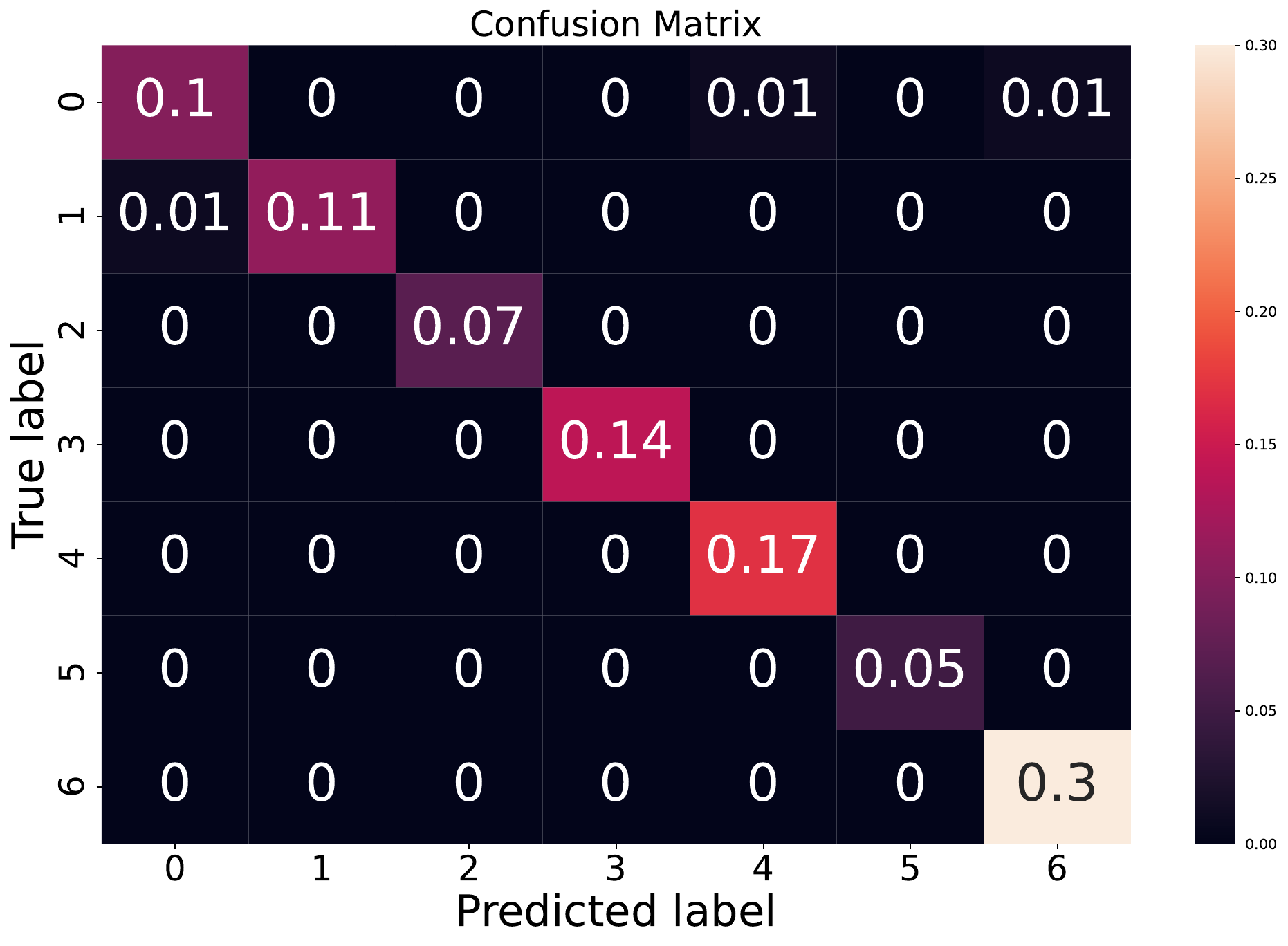}
    }

    \subfloat[Centralized MRI]{
        \includegraphics[width=0.48\textwidth]{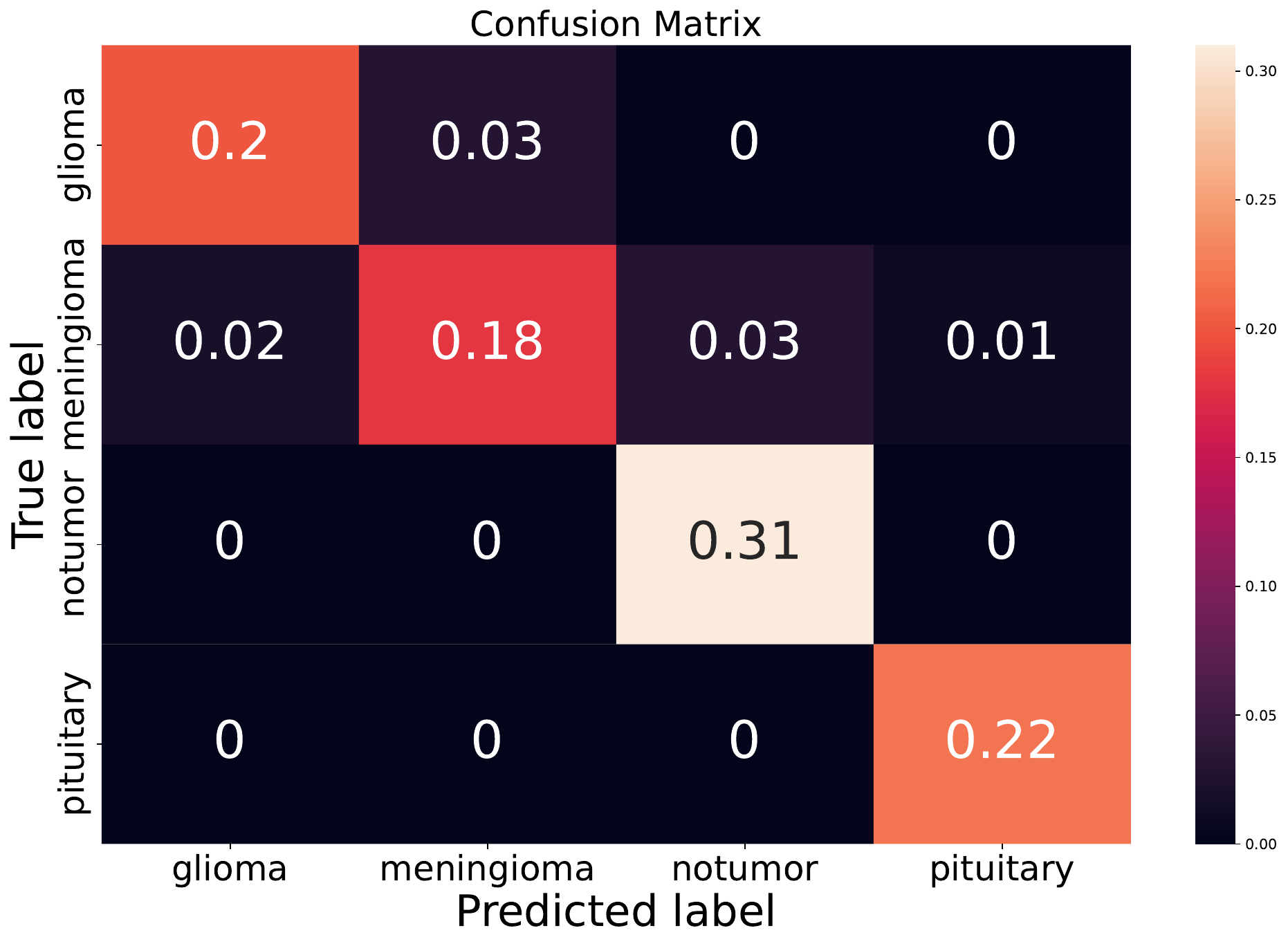}
    } 
    \hfill
    \subfloat[Federated MRI]{
        \includegraphics[width=0.48\textwidth]{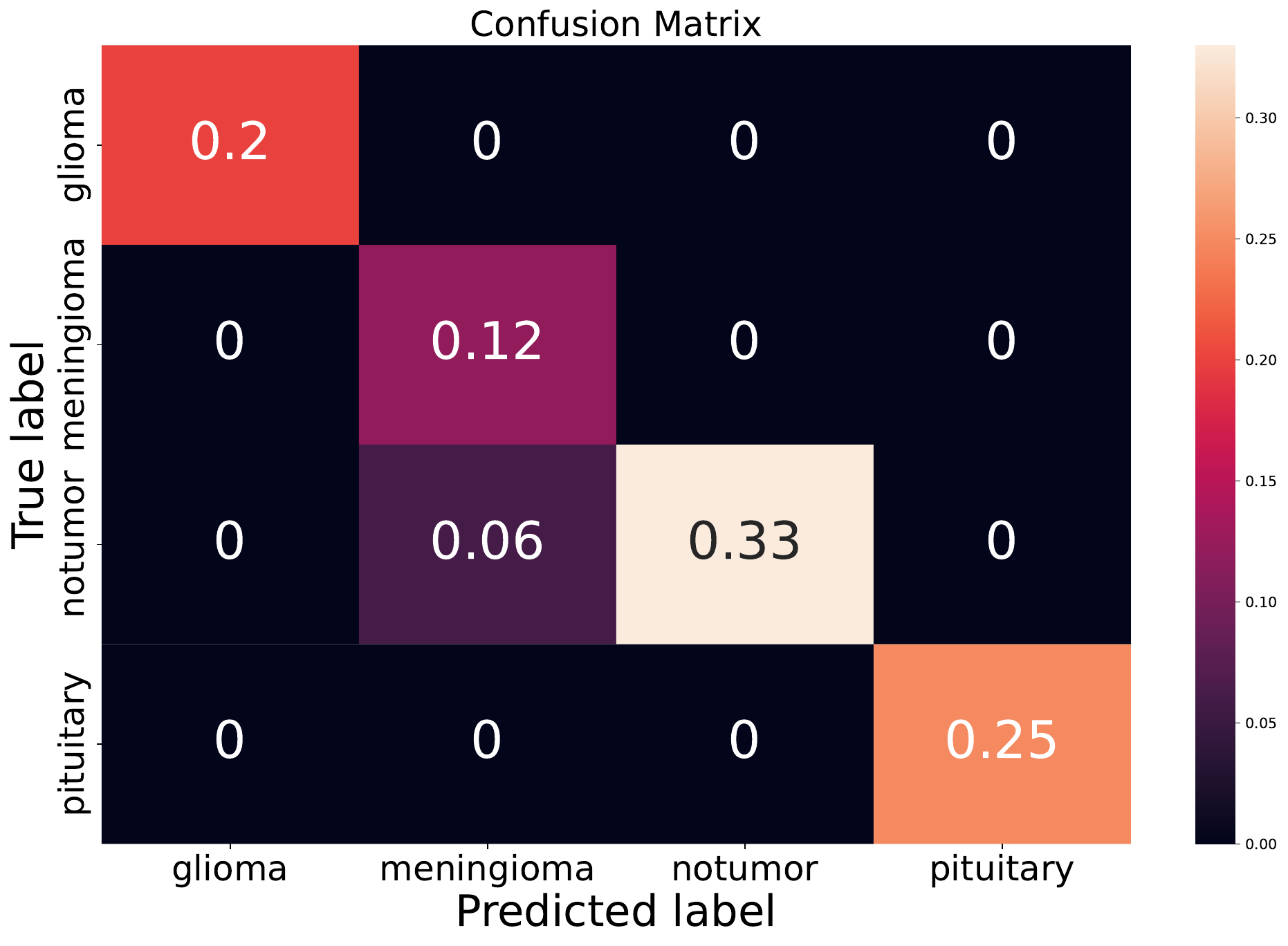}
    }
    \caption{Confusion matrix for the DNA+MRI MMoE.}
    \label{fig:mmoe_confusion_matrix}
\end{figure}

The confusion matrices for both the centralized and federated learning approach for each of the datasets, DNA and MRI, respectively, are given in Fig.~\ref{fig:mmoe_confusion_matrix}.
When comparing centralized and federated approaches for the DNA dataset, the federated model shows a notable improvement in classification performance and robustness.
Although the centralized approach shows a varied performance, with the highest classification accuracy being 31\% for class 6 and lower accuracies for the other classes (ranging from 7\% to 16\%), the federated model achieves similar accuracies for most classes, the differences being class 6 (30\%), class 3 (14\%) and class 4 (17\%).
The federated model also shows similar misclassification rates, as evidenced by the off-diagonal entries, indicating that both achieve proper class separation.
Although the accuracy for some classes, such as 0 and 1, decreases slightly in the federated model, this approach offers a consistent classification pattern across most classes, making it an effective strategy for handling the complexity of the DNA dataset.
When comparing centralized and federated approaches to the MRI dataset, the federated model improves classification accuracy and reduces misclassification for most classes.
The federated model increases the classification accuracy for the "notumor" class from 31\% to 33\% and for the "pituitary" class from 22\% to 25\% while maintaining the same accuracy for the "glioma" class at 20\%.
Additionally, the federated approach effectively eliminates certain misclassifications observed in the centralized model, such as the confusion between "glioma" and "meningioma", or "meningioma" and "notumor."
However, the "meningioma" class remains a challenging category for both models, with accuracies (18\% in the centralized model vs. 12\% in the federated model) indicating that further enhancements are needed to differentiate this class better.
The comparison between centralized and federated approaches for both MRI and DNA datasets demonstrates that the federated learning model within the MMoE framework consistently improves classification performance across different data modalities.
For MRI data, the federated model improves accuracy for key classes and reduces confusion between similar tumor types.
In contrast, for the DNA data, it shows better overall class separation and reduced misclassification rates.
These results highlight the advantage of the Multimodal Mixture of Experts in federated settings, as it effectively leverages diverse data sources to achieve more robust and accurate classification across heterogeneous datasets.
        
\begin{figure}[tbh]
    \centering
    \subfloat[Centralized DNA]{
        \includegraphics[width=0.48\textwidth]{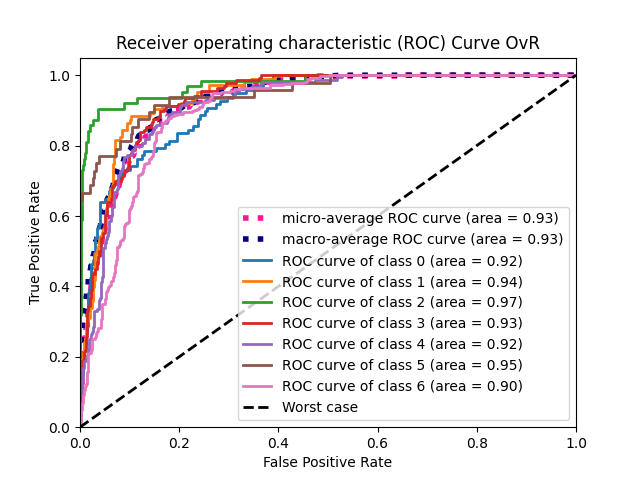}
    } 
    \hfill
    \subfloat[Federated DNA]{
        \includegraphics[width=0.48\textwidth]{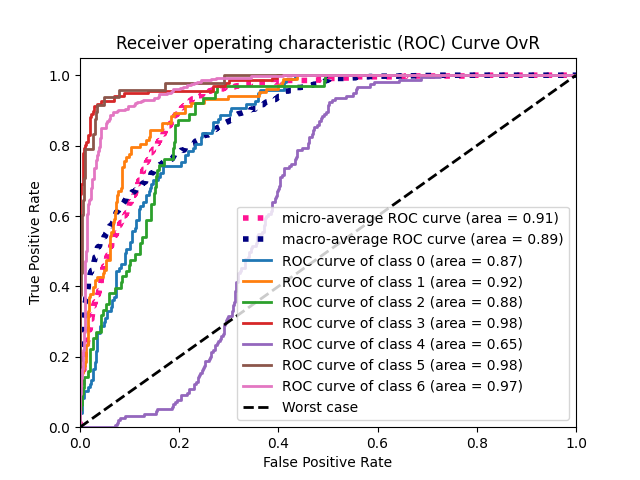}
    }

    \subfloat[Centralized MRI]{
        \includegraphics[width=0.48\textwidth]{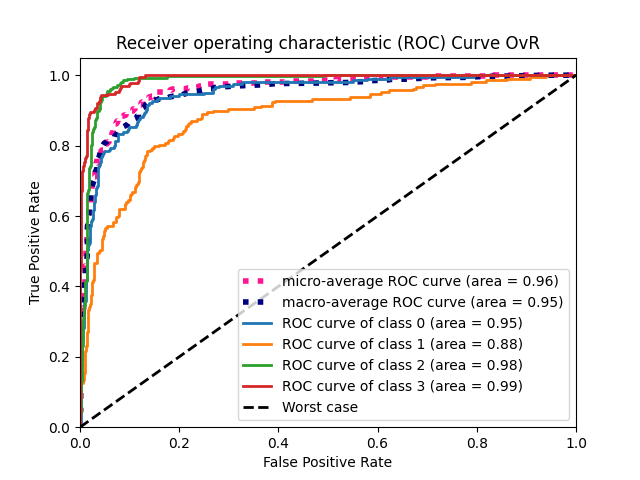}
    } 
    \hfill
    \subfloat[Federated MRI]{
        \includegraphics[width=0.48\textwidth]{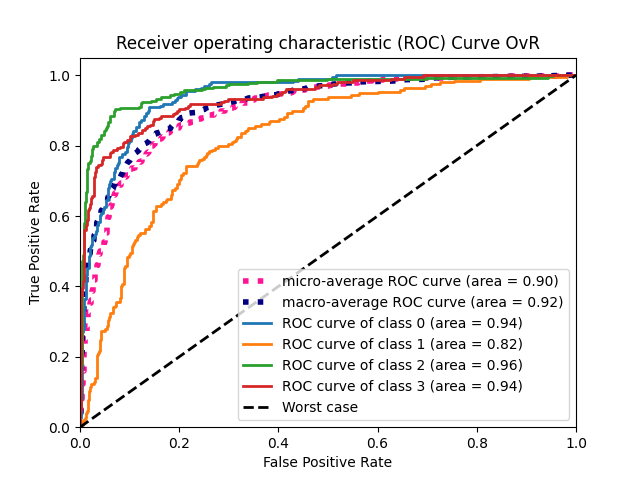}
    }
    \caption{Receiver operating characteristic curve (ROC) for DNA+MRI MMoE.}
    \label{fig:mmoe_roc}
\end{figure}

Similarly, we report the ROC curves for each dataset, DNA and MRI in the MMoE, both in the centralized and federated approach in Fig.~\ref{fig:mmoe_roc}.
When comparing centralized and federated models for the DNA dataset, they achieve a similar classification performance.
The micro-average ROC curves are 0.93 to 0.91, indicating high overall classification ability.
For example, class 0 only decreases from an area of 0.92 in the centralized model to 0.87 in the federated model.
In addition, classes 3, 5, and 6 achieve near-perfect ROC areas in the federated model, highlighting improved sensitivity and specificity and an advantage over the centralized approach.
Despite these gains, class 4 remains a challenging category with a relatively low ROC area of 0.65 in the federated model. This suggests that while federated learning improves performance in most classes, further optimization is needed for specific categories.
When comparing centralized and federated models for the MRI dataset, the centralized model demonstrates better performance across most metrics.
The micro-average and macro-average ROC curves for the centralized model achieve an area of 0.96 and 0.95, respectively, while the federated model drops to 0.90 and 0.92, respectively, indicating a decrease in overall classification ability.
Class-specific performance is also impacted in the federated model, with class 1 showing a significant drop in area from 0.88 to 0.82, highlighting its reduced capacity to accurately classify this class.
In contrast, the centralized model maintains high ROC areas for all classes, ranging from 0.88 to 0.99, showing that it can effectively capture distinguishing features across tumor types.
These results suggest that the centralized model is better suited to handle the MRI dataset, whereas the federated model struggles to integrate information from distributed sources, leading to a decline in classification performance for certain classes.

The comparison between centralized and federated models for both MRI and DNA datasets shows that the Multimodal Mixture of Experts framework can effectively leverage federated learning to enhance performance for complex datasets like DNA, where the federated model demonstrated good class-specific accuracy and consistent ROC curves.
While the federated model showed a slight decline in performance for the MRI dataset compared to the centralized approach, it still maintained strong classification capability for most classes, demonstrating its robustness in diverse data modalities.
These results highlight the advantage of the Multimodal Mixture of Experts in federated settings, as it allows for a tailored approach that optimally balances performance across different data types, ensuring that the model can adapt to varying data complexities while maintaining considerable generalization.

Table~\ref{tab:experiment-results} includes the results of the training and testing accuracies, training and testing losses, and training time for all experiments in this manuscript.
For completeness, we include the case where the DNA dataset is trained independently.
The comparison between the DNA-only and DNA+MRI rows in both centralized and federated cases reveals that incorporating MRI data in the Multimodal Mixture of Experts framework influences model performance across training configurations. In the centralized case, the addition of MRI data increases the test accuracy from 93.59\% for DNA alone to 95.20\% for DNA + MRI, and the test loss decreases from 0.8229 to 0.2925, indicating an improvement in the generalizability of the model.
In the federated setting, the inclusion of MRI data results in a slight increase in test accuracy from 94.09\% for DNA-only to 94.75\% for DNA in the multimodal DNA+MRI case, while the test loss decreases from 1.203 to 0.4167.
This suggests that while federated learning is effective in handling the increased complexity introduced by multimodal data, the integration of MRI information with DNA features poses challenges that can lead to a drop in performance.
Notably, in both federated and centralized cases, the training time increased significantly.
These observations indicate that the multimodal framework, while promising, requires sophisticated data fusion techniques to fully leverage the complementary nature of various datasets, in this case, DNA and MRI data, in both centralized and federated scenarios.

\subsection{Drug Discovery}

In this study, we adopt an FL framework to enhance the discovery of potential HIV inhibitors.
The primary focus of this work is to apply FL combined with graph neural networks (GNNs) created using \texttt{PyGeometric}~\cite{Fey/Lenssen/2019} to classify compounds as active or inactive against HIV.
This is based on the AIDS antiviral screening dataset from the Drug Therapeutics Program (DTP)\cite{wu2018moleculenet} found within the \texttt{MoleculeNet}, which uses representations of chemical compounds using SMILES (Simplified Molecular Input Line Entry System).

The dataset employed in this study evaluated the ability of more than 40,000 compounds to suppress HIV replication.
The results of this screening were categorized into three groups: confirmed inactive (CI), confirmed active (CA), and confirmed moderately active (CM).
The CA and CM groups are merged into a single "active" class, transforming the task into a binary classification problem: inactive (CI) versus active (CA + CM)~\cite{wu2018moleculenet}.

The model architecture used for this task is a Graph Convolutional Network (GCN)\cite{kipf2016semi}, which is well suited to learning from graph-structured data such as molecular compounds.
The model consists of an initial GCN layer followed by three additional convolutional layers.
Each convolutional layer is followed by a hyperbolic tangent activation function (tanh) to introduce non-linearity.
The output of the final layer is pooled using global min pooling (\texttt{gap}) and global max pooling (\texttt{GMP}) to capture both global features.

The performance of the FL-based GCN was evaluated using Root Mean Square Error (RMSE) loss and accuracy metrics.
The global model, which aggregates updates from clients in each round, was tested on a held-out validation set after each federated round.
In addition, client-specific evaluation metrics were recorded to assess model performance at each site. 
This graph-based approach maintains the privacy of participating institutions while enabling collaborative model development to accelerate the discovery of novel HIV inhibitors.

The confusion matrices for both centralized and federated learning cases in the GNN model for identifying promising molecules for HIV treatment show highly similar classification performance, with only minor differences.
In the centralized model, 95\% of the "CI" class (promising molecules) are correctly classified, while the federated model achieves a slightly higher accuracy of 96\% for the same class.
Both models correctly classify 2\% of the "CA/CM" class (non-promising molecules), with 1\% of "CA/CM" being misclassified as "CI" for federated and 4\% for centralized.
These results indicate that federated learning maintains a high classification accuracy comparable to the centralized model, particularly for the "CI" class, but neither approach effectively distinguishes the "CA/CM" class.
Due to space constraints, the confusion matrices are not included in the manuscript.

\begin{figure}[tbh]
    \centering
    \subfloat[Centralized]{
        \includegraphics[width=0.48\textwidth]{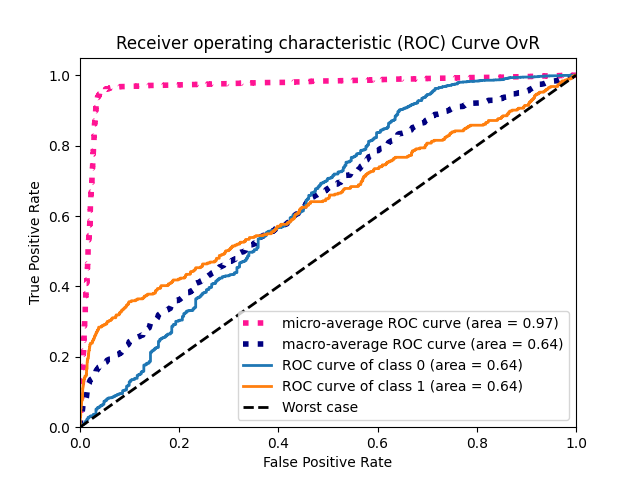}
    } 
    \hfill
    \subfloat[Federated]{
        \includegraphics[width=0.48\textwidth]{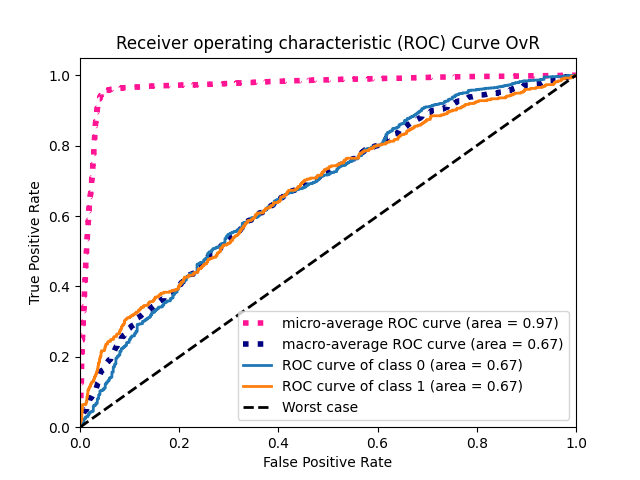}
    }
    \caption{Receiver operating characteristic curve (ROC) for the HIV dataset.}
    \label{fig:hiv_roc}
\end{figure}

The ROC curves for this case are presented in Fig.~\ref{fig:hiv_roc}.
Compared, centralized and federated models for the HIV dataset achieve strong overall classification performance, as indicated by the high micro-average ROC area of 0.97 in both cases.
However, the federated model shows a slight improvement in class-specific performance, with the ROC areas for the "CI" and "CA/CM" classes increasing from 0.64 in the centralized model to 0.67 in the federated model.
This improvement is also reflected in the macro-average ROC curve, which increases from 0.64 to 0.67.
These results suggest that the federated learning approach provides better differentiation between the two classes, possibly due to the model's ability to leverage diverse distributed data sources.
However, despite the incremental gains, both models still struggle to distinguish between "CI" and "CA/CM," indicating a need for further refinement to enhance classification specificity in the HIV dataset.

The comparisons between the centralized and federated models for the DNA and MRI datasets in the Multimodal Mixture of Experts (MMoE) framework and the HIV dataset in the GNN framework demonstrate the potential of federated learning for chemical engineering applications.
For the DNA dataset, federated learning improved class-specific performance and enhanced the model's overall capability, particularly in handling complex, heterogeneous data.
The federated model showed good class separation and reduced misclassification, resulting in similar ROC areas when compared to the centralized model.
Similarly, while the federated model for the MRI dataset did not surpass the high performance of the centralized model, it still maintained a substantial classification accuracy for most classes.
This result suggests that federated learning can effectively handle multimodal and complex datasets often found in chemical engineering applications, where data privacy and collaboration between different organizations are critical.

In the HIV dataset, where a GNN was used to classify promising molecules for treatment, federated learning demonstrated a slight improvement in class-specific and macro-average ROC areas compared to the centralized model.
Although the federated model did not fully resolve the challenge of distinguishing between the "CI" and "CA/CM" classes, it showed better consistency across distributed data sources and maintained a high overall classification performance.
This reinforces the potential of federated learning to enhance model robustness and generalization when working with diverse chemical datasets, such as those encountered in drug discovery or material design.
Federated learning can be a powerful tool for chemical engineering domains by preserving data privacy while enabling collaborative model training.
It facilitates the development of predictive models that can draw from a wider variety of data sources without compromising data security or proprietary information.

\section{Results Summary}
We summarize the performance of the different examples in this manuscript in Table~\ref{tab:experiment-results}.
The table presents the training and test accuracies, training and test losses, and execution times for centralized and federated experiments conducted on the PILL, DNA, MRI, and HIV datasets. It includes results from 200 epochs of centralized training for one round and 10 epochs of federated training over 20 rounds with 10 clients present for each round for a comparative analysis of model performance across different configurations.
The table results highlight the differences in model performance between centralized and federated learning configurations across the PILL, DNA, MRI, and HIV datasets.
Overall, federated models demonstrated comparable generalization capabilities for the PILL and DNA datasets, with improved test accuracies and lower test losses, indicating that federated learning effectively maintains performance even in distributed scenarios.
However, the addition of MRI data in the multimodal DNA+MRI case led to a noticeable drop in test accuracy, suggesting that handling heterogeneous data modalities remains a challenge.
Furthermore, federated training incurred comparable computational costs in terms of training time across all datasets, and it was even faster for some datasets, showing that it is possible to improve privacy without sacrificing computational complexity.
Nevertheless, these findings suggest that while federated learning holds promise for chemical engineering applications where data sharing and privacy are critical, further optimization is needed to better improve performance and computational efficiency, especially when dealing with multimodal data sources.
All the implementations that led to the results presented in this manuscript are openly available in the repository \href{https://github.com/elucidator8918/Fed-ML-Chem}{https://github.com/elucidator8918/Fed-ML-Chem}.

\begin{table}[tbh]
\centering
\renewcommand{\arraystretch}{1.4}
\begin{tabular}{@{}lrrrrr@{}}
\toprule
\textbf{Experiment} & \textbf{Train Acc} & \textbf{Test Acc} & \textbf{Train Loss} & \textbf{Test Loss} & \textbf{Time (sec)} \\
\midrule
CL PILL & 99.69\% & 93.27\% & 0.0204 & 0.1576 & 3019.34 \\
FL PILL & 93.54\% & 94.79\% & 0.2800 & 0.207 & 2215.40 \\
CL DNA & 100.00\% & 93.59\% & 0.0000 & 0.8229 & 1343.17 \\
FL DNA & 100.00\% & 94.09\% & 0.0000 & 1.203 & 3921.43 \\
CL DNA (MMoE) & 100.00\% & 95.20\% & 0.0000 & 0.2925 & \multirow{2}{*}{6173.90} \\
CL MRI (MMoE) & 100.00\% & 90.88\% & 0.0000 & 1.2997 & \\
FL DNA (MMoE) & 99.00\%   & 94.75\% & 0.0776 & 0.4167 & \multirow{2}{*}{5543.29} \\
FL MRI (MMoE) & 99.38\%  & 85.56\% & 0.1997 & 1.0720 & \\
CL HIV & 97.15\% & 95.51\% & 0.1520 & 0.1804 & 1716.63 \\
FL HIV & 96.31\% & 95.34\% & 0.1790 & 0.1870 & 1042.82 \\
\bottomrule
\end{tabular}
\caption{Results of Centralized (CL) and Federated (FL) Experiments. In FL, both train loss and accuracy refer to a single client. The MMoE experiments share the training time column because they are trained as a single model.}
\label{tab:experiment-results}
\end{table}

\section{Future Aspects of Federated Learning and Perspectives}

When applied in an industrial setting, FL faces the challenge of maintaining privacy and data confidentiality also during communication with the server~\cite{zellinger_beyond_2021}.
A collaborative context where many manufacturers send their models for aggregation poses the risk of external attackers, or even other clients in the same training scheme, being able to reconstruct the data from the shared information.
This lack of trust in the environment suggests a need for better guarantees regarding data privacy during aggregation.

Recent work has focused on providing the tools to guarantee data confidentiality while transmitting the models through a network.
In this section, we briefly present promising approaches for tackling these obstacles.

\subsection{Working with Encrypted Data}

In FL, the client data remains local, with only model updates distributed throughout the network.
However, these models may still contain important information that external attackers could try to exploit.
A potential solution is to encrypt the models before transmitting them~\cite{fang2021privacy}.
With \emph{Fully Homomorphic Encryption} (FHE)~\cite{gentry2009fhe}, the server can perform the aggregation step directly on encrypted models.
Although FHE incurs computational overhead, recent work has focused on improving efficiency to mitigate this drawback~\cite{xie2024efficiency, rieyan2024advanced}.

\subsection{Blockchain}

Several proposals have emerged to address these security concerns, such as implementing more robust verification techniques or adopting decentralized trust mechanisms.\cite{SMarteen}
One such proposal is the use of consensus algorithms to validate model updates, ensuring that malicious activities are identified before they impact the global model.
Another involves periodically auditing the model's updates to ensure that they meet specific integrity standards.
However, these methods still depend on a centralized authority or require complex cryptographic solutions that can introduce computational overhead.\cite{Cabaj2018-1}

Despite these efforts, existing proposals often fail to address the scalability and complexity of real-world FL systems.
For example, consensus-based approaches can become computationally expensive when scaled across many devices.
Likewise, centralized audit mechanisms remain vulnerable to attacks targeting the central server, while cryptographic solutions may not be practical for resource-constrained devices that often participate in FL, such as smartphones or IoT devices.

Blockchain technology presents a promising solution to these challenges by providing a decentralized, tamper-resistant ledger that can securely track and verify model updates in FL.
By integrating blockchain, each update can be recorded in an immutable ledger that is shared among all participants, ensuring that any manipulation or malicious activity is immediately detectable.
Blockchain’s consensus mechanisms, such as proof of work or proof of stake, can help ensure the validity of updates and build trust without relying on a centralized server.

Some research, such as the one conducted by \citet{VMothukuri}, seeks to highlight vulnerabilities to be considered in different architectures, including the possibility of model poisoning through tampering with historical records.

Developing a secure historical record poses challenges, especially in the context of distributed systems.
Since it depends on communication between clients and servers, this communication is subject to failures or malicious attacks and, therefore, needs to provide auditability when necessary and, in some instances, introduce some level of difficulty for potential attackers.
Some works aim to present solutions to this problem.
\citet{BXianglin}, in proposing \textit{FLChain}, introduces the concept of a global model stored in a Blockchain structure similar to that implemented in the cryptocurrency \textit{Ethereum}.
The authors suggest using consensus algorithms to update the global model and generate new blocks, with clients communicating directly with each other during the process.
This solution presents particular challenges for implementation, as it depends on a structure without a central aggregator.

Similarly, \textit{BlockAudit}, as proposed by \citet{AhmadA}, suggests an approach developed for systems without a central server, where an external blockchain network is proposed to allow transactions and consensus between clients.
Like the previous one, it presents challenges and requires the system to adapt to the proposed architecture.
\citet{ChainFL} presents the concept of \textit{BlockFL}, which uses a distributed block generation and reward system as a basis for implementing a distributed FL system.

\subsection{Federated Ensemble Learning}

Federated ensemble learning is an advanced approach that combines the principles of ensemble learning with FL to further enhance the performance of the model in decentralized settings \cite{chen2020fedbe}.
Instead of training a single global model across distributed clients, federated ensemble learning allows each client to train its own model locally. The central server aggregates the predictions of these diverse models to form an ensemble.
This approach takes advantage of the natural diversity of local models, which can be optimized for different subsets of the data, leading to improved generalization and robustness compared to a single global model.
Federated ensemble learning can be particularly effective when clients have heterogeneous data distributions, as each local model can capture unique patterns within its specific dataset.
By aggregating these models, the system benefits from their collective wisdom without requiring direct access to client data, thereby maintaining FL's privacy-preserving advantages \cite{subashchandrabose2023ensemble,zhao2023ensemble}.

\subsection{TinyML}

\emph{TinyML} and FL represent a powerful combination to enable intelligent decentralized processing in resource-constrained environments.\cite{ficco2024federated}
TinyML focuses on the deployment of ML models on edge devices with limited computational power, such as microcontrollers and IoT sensors, allowing real-time data analysis directly on the device.
FL complements this by enabling these devices to collaboratively train models without sharing raw data, preserving privacy, and reducing the need for constant communication with a central server.\cite{adhikary2024fedtinywolf}
By combining TinyML and FL, it is possible to create highly efficient and scalable AI systems that can operate autonomously in environments where connectivity is intermittent, or data privacy is paramount, such as wearable healthcare devices, industrial IoT systems, and smart home devices. \cite{qi2024small}
This integration promotes privacy-conscious intelligence at the edge, reducing latency and energy consumption while improving security.\cite{lu2024lyapunov}

\subsection{Quantum Federated Learning}

Due to recent advancements in quantum encryption protocols~\cite{Pirandola:20}, another way to enhance security in a federated setting is to incorporate quantum computing layers into the distributed client's architecture~\cite{chen_federated_2021}.
\emph{Quantum federated learning} (QFL) builds on the strengths of \emph{Quantum machine learning} (QML)~\cite{biamonte2017quantum, rebentrost2014quantum} to tackle security-related issues found in decentralized data~\cite{huang2022quantum}.

Although QFL holds great promise, it is still an emerging technology and faces challenges in the implementation and management of resources~\cite{larasati2022QuantumFederatedLearning, ren2023towards}.
However, for some systems, the quantum layers of QFL can be simulated on classical hardware using tensor networks~\cite{bhatia2024federatedhierarchicaltensornetworks}.

Recent efforts explore the integration between QFL and encrypted ML models~\cite{chu2023cryptoqfl,zhang2022federated},
including the use of fully homomorphic encryption~\cite{dutta2024federatedlearningquantumcomputing}~\cite{dutta2024mqfl}

\subsection{Perspectives}

As presented in this manuscript, Federated Learning has the potential to be a valuable and impactful tool for implementing artificial intelligence in settings relevant to chemical engineering.
Various applications are very interested in the power of utilizing distributed computing to achieve a common learning goal while providing privacy guarantees for each local client.
Through this work, we aim to present this tool to a community that we believe will significantly benefit from it.
In many other research areas, chemical engineering has seen an increase in the use of ML tools.
This, combined with some privacy constraints associated with some cases, provides fertile ground for FL to become one more tool for tackling the increasingly complex challenges ahead.
We exemplify how this distributed setting can take advantage of successful centralized methods, such as deep, convolutional, and graph neural networks, and can even integrate different data inputs.
Although we acknowledge the limitations, following the title of the work by \citet{zellinger_beyond_2021}, we consider this to be a first step toward considering what lies beyond federated learning, particularly with applications in chemical and process engineering.

\newpage

\begin{suppinfo}

This supplementary material provides code snippets used to implement FL using the \texttt{Flower} and \texttt{TensorFlow Federated} frameworks.
Consider that these snippets are taken from our repository \href{https://github.com/elucidator8918/Fed-ML-Chem}{https://github.com/elucidator8918/Fed-ML-Chem} at the time of submission and work with the following software in these particular versions: \texttt{Python 3.10}, \texttt{flower 1.5.0}, and \texttt{tensorflow-federated 0.86.0}.

\subsection{Flower Implementation}
\begin{figure}
\caption*{Code Snippet 1: Custom Strategy Definition (e.g., \textit{FedAvg}) using \texttt{Flower}.}
\begin{minted}{python}
|{\color{customblue}strategy}| =|{\color{customred} FedCustom}|(
    fraction_fit=frac_fit,
    fraction_evaluate=frac_eval,
    min_fit_clients=min_fit_clients,
    min_evaluate_clients=min_eval_clients if min_eval_clients else number_clients // 2,
    min_available_clients=min_avail_clients,
    evaluate_metrics_aggregation_fn=weighted_average,
    initial_parameters=ndarrays_to_parameters(get_parameters(central)),
    evaluate_fn=evaluate,
    on_fit_config_fn=get_on_fit_config_fn(epoch=max_epochs, batch_size=batch_size),
)
\end{minted}
\end{figure}

\begin{figure}
\caption*{Code Snippet 2: Initiation of the Simulation in a Federated Environment using \texttt{Flower}.}
\begin{minted}{python}
|{\color{BlueGreen}fl.simulation.start\_simulation}|(
    client_fn=client_fn,
    num_clients=number_clients,
    config=fl.server.ServerConfig(num_rounds=rounds),
    strategy =|{\color{customblue} strategy}|,
    client_resources=client_resources
)
\end{minted}
\end{figure}
\FloatBarrier
\subsection{TensorFlow Federated Implementation}
\begin{figure}
\caption*{Code Snippet 3: Definition of a Keras Model and Conversion to a \texttt{TensorFlow Federated} Model.}
\begin{minted}{python}
def model_fn():
    model = PillModel(image_shape, num_categories)
    return tff.learning.models.from_keras_model(
        keras_model=model.model,
        input_spec=model.input_spec,
        loss=tf.keras.losses.SparseCategoricalCrossentropy(),
        metrics=[tf.keras.metrics.CategoricalAccuracy()]
    )
\end{minted}
\end{figure}

\begin{figure}
\caption*{Code Snippet 4: Federated Training of the Model Using a Distributed Dataset using \texttt{TensorFlow Federated}.}
\begin{minted}{python}
def federated_train(data_folder, num_rounds):
    dataset_train = load_data(data_folder + "/Training", num_clients)
    dataset_train.preprocess(lambda ds: ds.shuffle(shuffle_buffer).repeat(num_epochs))

    process = tff.learning.algorithms.build_weighted_fed_avg(
        model_fn=model_fn,
        client_optimizer_fn=optimizer_fn,
        server_optimizer_fn=optimizer_fn
    )
    state = process.initialize()

    for round in range(num_rounds):
        selected_clients = np.random.choice(dataset_train.client_ids, size=num_clients, replace=False)
        federated_train_data = [dataset_train.create_tf_dataset_for_client(x) for x in selected_clients]
        state, metrics = process.next(state, federated_train_data)

    return process, state
\end{minted}
\end{figure}

\end{suppinfo}

\FloatBarrier

\bibliography{main}

\end{document}